\documentclass[letterpaper]{article} 
\usepackage{aaai25}  
\usepackage{times}  
\usepackage{helvet}  
\usepackage{courier}  
\usepackage[hyphens]{url}  
\usepackage{graphicx} 
\usepackage{natbib}  
\usepackage{caption} 
\usepackage{amsmath}
\usepackage{amssymb}
\usepackage{amsthm}
\usepackage[linesnumbered,ruled,vlined,noend]{algorithm2e}
\usepackage{booktabs}
\usepackage{siunitx}
\usepackage{cleveref}
\usepackage{multirow}
\usepackage{adjustbox}
\usepackage{pifont}
\usepackage{dsfont}
\usepackage{subcaption}
\usepackage{enumitem}
\usepackage{placeins}

\Crefname{figure}{Fig.}{Figs.}
\Crefname{equation}{Eq.}{Eqs.}

\DeclareMathOperator{\logonep}{log1p}

\newtheoremstyle{my_definition}
  {\topsep} 
  {\topsep} 
  {\itshape} 
  {0pt} 
  {\bfseries\itshape} 
  {\textmd{.}} 
  { } 
  {\textup{\thmname{#1}\thmnumber{ #2}:} \thmnote{#3}} 

\theoremstyle{my_definition}
\newtheorem{definition}{Definition}

\title{Holistic Semantic Representation for Navigational Trajectory Generation}
\author{
    Ji Cao\textsuperscript{\rm 1},
    Tongya Zheng\textsuperscript{\rm 2,3,}\thanks{Corresponding author.},
    Qinghong Guo\textsuperscript{\rm 1},
    Yu Wang\textsuperscript{\rm 1},
    Junshu Dai\textsuperscript{\rm 1},\\
    Shunyu Liu\textsuperscript{\rm 4},
    Jie Yang\textsuperscript{\rm 1},
    Jie Song\textsuperscript{\rm 1},
    Mingli Song\textsuperscript{\rm 3,5}
}
\affiliations{
    \textsuperscript{\rm 1}Zhejiang University\\
    \textsuperscript{\rm 2}Big Graph Center, Hangzhou City University\\
    \textsuperscript{\rm 3}State Key Laboratory of Blockchain and Data Security, Zhejiang University\\
    \textsuperscript{\rm 4}Nanyang Technological Univerisity\\
    \textsuperscript{\rm 5}Hangzhou High-Tech Zone (Binjiang) Institute of Blockchain and Data Security\\
    \{caoji2001, q.h\_guo, yu.wang, djs, yang\_jie, sjie, brooksong\}@zju.edu.cn,\\
    doujiang\_zheng@163.com, shunyu.liu@ntu.edu.sg
}

\begin{document}

\maketitle

\begin{abstract}
Trajectory generation has garnered significant attention from researchers in the field of spatio-temporal analysis, as it can generate substantial synthesized human mobility trajectories that enhance user privacy and alleviate data scarcity. However, existing trajectory generation methods often focus on improving trajectory generation quality from a singular perspective, lacking a comprehensive semantic understanding across various scales. Consequently, we are inspired to develop a \textbf{HO}listic \textbf{SE}mantic \textbf{R}epresentation (HOSER) framework for navigational trajectory generation. Given an origin-and-destination (OD) pair and the starting time point of a latent trajectory, we first propose a Road Network Encoder to expand the receptive field of road- and zone-level semantics. Second, we design a Multi-Granularity Trajectory Encoder to integrate the spatio-temporal semantics of the generated trajectory at both the point and trajectory levels. Finally, we employ a Destination-Oriented Navigator to seamlessly integrate destination-oriented guidance. Extensive experiments on three real-world datasets demonstrate that HOSER outperforms \emph{state-of-the-art} baselines by a significant margin. Moreover, the model's performance in few-shot learning and zero-shot learning scenarios further verifies the effectiveness of our holistic semantic representation.
\end{abstract}

\begin{links}
\link{Code}{https://github.com/caoji2001/HOSER}
\end{links}

\section{Introduction}

With the rapid development of Global Positioning Systems (GPS) and Geographic Information Systems (GIS), the number of human mobility trajectories has soared, significantly advancing research in spatio-temporal data mining, such as urban planning~\cite{bao2017planning, wang2023human, wang2024cola, wang2024star}, business location selection~\cite{li2016mining}, and travel time estimation~\cite{reich2019survey, wen2024survey}. However, due to obstacles including privacy issues~\cite{cao2021generating}, government regulations~\cite{chen2024deep}, and data processing costs~\cite{zheng2015trajectory}, it is not easy for researchers to obtain high-quality real-world trajectory data. A promising solution to these challenges is trajectory generation, which not only meets privacy requirements but also allows for the creation of diverse high-fidelity trajectories. These trajectories are capable of producing similar data-analysis results, supporting broader research and application needs.

In addition to traditional statistical methods~\cite{song2010modelling, jiang2016timegeo}, deep learning has improved trajectory generation by encoding fine-grained human mobility semantics in high-dimensional representations. A series of trajectory generation methods employ RNNs and CNNs to capture spatio-temporal features in the trajectories, along with various generative models such as VAEs~\cite{huang2019variational, lestyan2022search}, GANs~\cite{cao2021generating, wang2021large}, and diffusion models~\cite{zhu2023difftraj}. In addition, another line of methods incorporates the connectivity of spatio-temporal points by embedding the topological semantics of road networks into trajectory generation~\cite{feng2020learning, wang2024stega, zhu2024controltraj}. However, since experienced drivers~\cite{yuan2010tdrive} often identify the fastest spatio-temporal routes to their destinations, previous methods have substantially overlooked the impact of destination on generated trajectories, resulting in a deviation from practical realities.

To the best of our knowledge, only TS-TrajGen~\cite{jiang2023continuous} incorporates both origin and destination locations in trajectory generation based on the A* algorithm. However, TS-TrajGen strictly adheres to the principle of the A* algorithm to separately model the semantics of the trajectory level and the road level in a two-tower paradigm, which hinders semantic sharing and end-to-end learning in trajectory generation. In general, existing methods lack a comprehensive understanding of the relationships between road segments, trajectories, and their origins and destinations.

Therefore, we are motivated by the semantic relationships to develop a \textbf{HO}listic \textbf{SE}mantic \textbf{R}epresentation (HOSER) framework for navigational trajectory generation. Using a bottom-up approach, we first derive long-range road semantics by partitioning road networks into a hierarchical topology. The trajectory representations are then encoded in a multi-granularity manner to integrate spatio-temporal dynamics with road-level semantics. Finally, we guide the trajectory generation process by incorporating both the semantic context of partial trajectories and the semantics of the destination. During the generation phase, HOSER iteratively predicts the probabilities of candidate road segments based on a progressively generated trajectory and its destination. Extensive experimental results and visualization analyses on three real-world trajectory datasets demonstrate that our proposed HOSER framework achieves significantly better trajectory generation quality than \emph{state-of-the-art} baselines in both global and local level metrics. Furthermore, these generated trajectories can be effectively applied to downstream tasks, demonstrating their great potential to replace real trajectories for spatio-temporal data analysis. In addition, due to its outstanding architectural design, HOSER demonstrates exceptional performance in few-shot and zero-shot learning scenarios. In summary, our contributions can be summarized as follows:

\begin{itemize}
\item We identify a significant representation gap among road segments, trajectories, and their respective origins and destinations in trajectory generation, which is frequently overlooked by existing trajectory generation methods.
\item We propose a novel \textbf{HO}listic \textbf{SE}mantic \textbf{R}epresentation (HOSER) framework, which is designed to bridge the aforementioned semantic gap in trajectory generation by holistically modeling human mobility patterns.
\item We validate HOSER on three real-world trajectory datasets, demonstrating its ability to generate high-fidelity trajectories that surpass baselines at both the global and local levels. Furthermore, HOSER achieves satisfactory results in few-shot and zero-shot learning.
\end{itemize}

\section{Preliminary}

\begin{definition}[Road Network]
The road network is represented as a directed graph $\mathcal{G} = \langle \mathcal{V}, \mathcal{E} \rangle$, where $\mathcal{V}$ denotes the set of road segments (nodes), and $\mathcal{E}$ denotes the set of intersections (edges) between adjacent road segments.

\textup{Note that road segments are defined as nodes rather than edges, following the widely adopted settings in previous studies~\cite{jepsen2019graph, wu2020learning}.}
\end{definition}

\begin{definition}[Trajectory]
We denote a trajectory as a sequence of spatio-temporal points $\tau = [x_1, x_2, \ldots, x_n]$. Each spatio-temporal point is represented as $x_i = (r_i, t_i)$, which is a pair of road segment ID and timestamp. The sequence ensures that each road segment $r_i$ is reachable from the previous segment $r_{i-1}$ for all $i \in [2, n]$.

\textup{Note that not all adjacent segments are reachable due to the prescribed driving direction on each road segment.}
\end{definition}

\begin{definition}[Trajectory Generation]
Given a set of real-world trajectories $\mathcal{T} = \{\tau^1, \tau^2, \ldots, \tau^m\}$, the objective of our trajectory generation task is to learn a $\theta$-parameterized generative model $G_\theta$. When given a triplet containing the origin road segment, the departure time, and the destination road segment $(r_\text{org}, t_\text{org}, r_\text{dest})$ as conditions, model $G_\theta$ is capable of generating a synthetic trajectory $[x_1, x_2, \ldots, x_n]$ such that $x_1 = (r_\text{org}, t_\text{org})$, and $r_n = r_\text{dest}$.
\end{definition}

\begin{definition}[Human Movement Modeling]
We approach the problem of generating high-quality trajectories by modeling the human movement policy $\pi(a | s)$, which gives the probability of taking action $a$ given the state $s$. Here, state $s$ includes the current partial trajectory $x_{1:i} = [x_1, \ldots, x_i]$ and the destination $r_\text{dest}$, action $a$ denotes moving to a currently reachable road segment, which can be written as:
\begin{equation}
\pi(a | s) = P(r_{i+1} | x_{1:i}, r_\text{dest}).
\end{equation}
Then the generation process can be seen as searching for the optimal trajectory with the maximum probability:
\begin{equation}
\resizebox{.875\columnwidth}{!}{$
\begin{aligned}
\max \prod_{i=1}^{n-1}\pi(a_i, s_i) &= \max \prod_{i=1}^{n-1}P(r_{i+1} \mid x_{1:i}, r_\text{dest}),\\
\mathrm{s.t.} \quad x_1 &= (r_\text{org}, t_\text{org}), \quad r_n = r_\text{dest}.
\end{aligned}
$}
\label{eq:trajectory_with_the_maximum_probability}
\end{equation}
Our task is to use neural networks to estimate the movement strategy $P_\theta(r_{i+1} | x_{1:i}, r_\text{dest})$ and the corresponding timestamp $t_{i+1}$ for the next spatio-temporal point.
\end{definition}

\section{Methodology}

\begin{figure*}[t]
\centering
\includegraphics[width=1.0\textwidth]{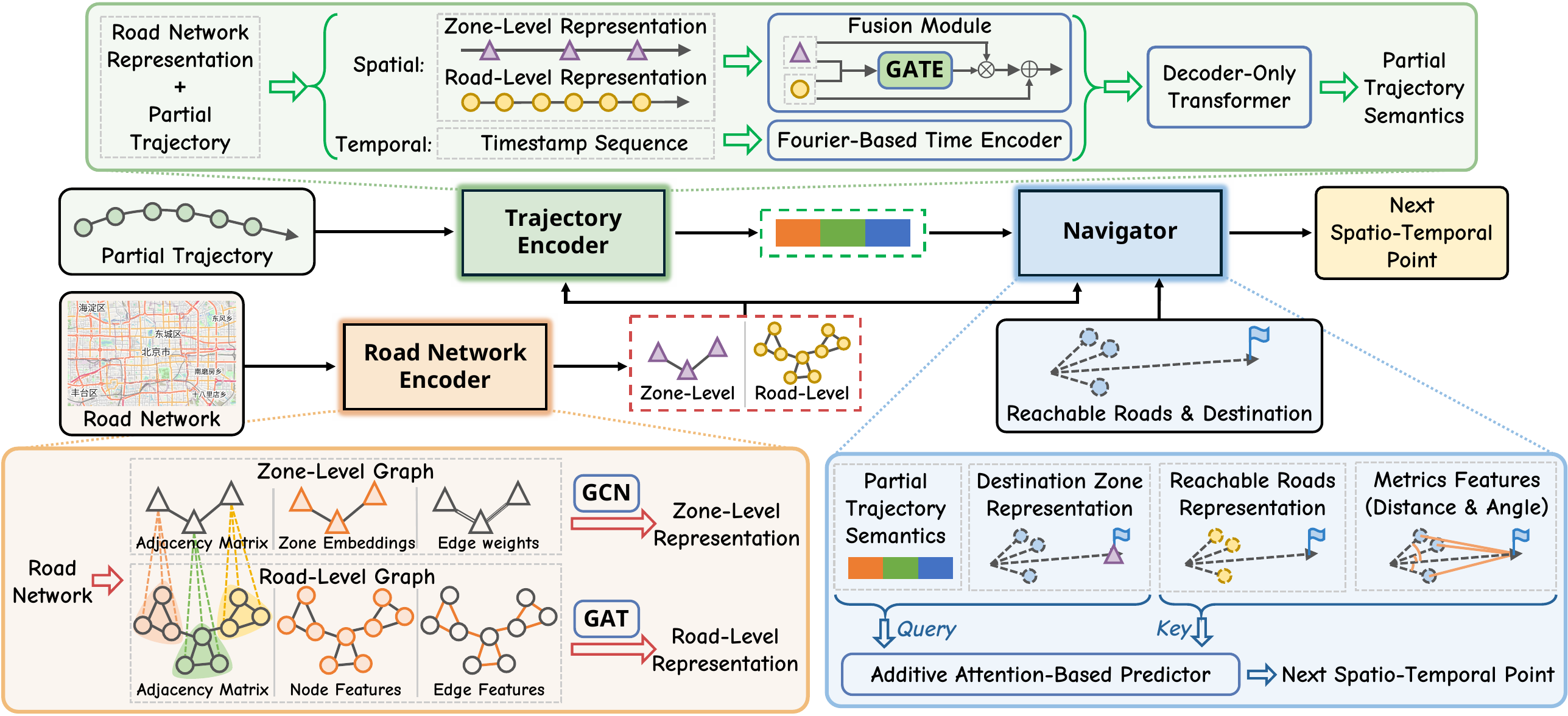}
\caption{The overview of the proposed HOSER framework. The Road Network Encoder is responsible for modeling the road network at different levels. The Trajectory Encoder is used to extract semantic information from the partial trajectory, which is then fed into the Navigator and combined with destination guidance to generate the next spatio-temporal point.}
\label{fig:framework}
\end{figure*}

In this section, we detail the proposed HOSER framework, which predicts the next spatio-temporal point based on the current state and generates the trajectory between the given OD pair through a search-based method. As illustrated in \Cref{fig:framework}, HOSER first employs a Road Network Encoder to model the road network at different levels. Based on the road network representation, a Multi-Granularity Trajectory Encoder is proposed to extract the semantic information from the current partial trajectory. To better incorporate prior knowledge of human mobility, a Destination-Oriented Navigator is used to seamlessly integrate the current partial trajectory semantics with the destination guidance.

\subsection{Road Network Encoder}

The road network is a fundamental part of the transportation system, and accurately modeling it is crucial for generating high-quality trajectories. However, designing effective representation learning methods remains challenging~\cite{han2021graph}. On the one hand, the road network's inherent topological structure means that connected road segments often correlate; on the other hand, non-connected road segments can still exhibit functional similarities, such as belonging to the same commercial or residential zone. Inspired by HRNR~\cite{wu2020learning}, we model the road network at both the road segment and zone levels to better capture the long-distance dependencies between road segments. Additionally, we use a deterministic road segment-to-zone allocation mechanism, which simplifies the complex allocation matrix learning process seen in HRNR~\cite{wu2020learning}.

\paragraph{Road-Level Semantic Representation.}

As outlined in Definition 1, we represent the road segments in the road network as nodes in the graph, with intersections between adjacent roads depicted as edges. The Road Network Encoder then encodes the road segments and intersections separately.

For the $i$-th road segment in the road network, we encode its road segment ID and its attributes (comprising four kinds: length, type, longitude, and latitude). These encoded attributes are then concatenated to form the road segment embedding $\boldsymbol{v}_i \in \mathbb{R}^d$, which can be written as $\boldsymbol{v}_i = \boldsymbol{v}_{\text{ID}} \ \Vert \ \boldsymbol{v}_{\text{len}} \ \Vert \ \boldsymbol{v}_{\text{type}} \ \Vert \ \boldsymbol{v}_{\text{lon}} \ \Vert \ \boldsymbol{v}_{\text{lat}}$, where $\boldsymbol{v}_{(\cdot)}$ denotes the embedding vector for a certain type of the road segment feature and ``$\Vert$'' is the concatenation operation.

For the intersection between road segments $i$ and $j$, we strengthen the directed road network by bidirected intersection embedding $\boldsymbol{e}_{ij} \in \mathbb{R}^d$, concatenating various features as $\boldsymbol{e}_{ij} = \mathds{1}_{ij} \ \Vert \ \boldsymbol{\phi}_{ij}$, where $\mathds{1}_{ij}$ and $\boldsymbol{\phi}_{ij}$ denote the embeddings for reachability and steering angle, respectively.

Then we employ GATv2~\cite{brody2022attentive} to fuse the aforementioned contextual embeddings of road segments and intersections, obtaining representations for the road segments within the road network at the $(\ell + 1)$-th layer:
\begin{equation}
\boldsymbol{v}_i^{(\ell+1)} = \sum_{j \in \mathcal{N}(i) \cup \{i\}} \alpha_{ij}^{(\ell)} \boldsymbol{v}_j^{(\ell)} \boldsymbol{\Theta}_t,
\end{equation}
where the attention coefficients $\alpha_{ij}^{(\ell)}$ are computed as:
\begin{equation}
\alpha_{ij}^{(\ell)} = \textrm{Softmax} \big( \sigma( \boldsymbol{v}_i \boldsymbol{\Theta}_s^{(\ell)} + \boldsymbol{v}_j \boldsymbol{\Theta}_t^{(\ell)} + \boldsymbol{e}_{ij} ) (\boldsymbol{a}^{(\ell)})^{\top} \big),
\end{equation}
which incorporates both road- and intersection-aware semantics. Here, $\sigma$ represents the LeakyReLU activation function, $\boldsymbol{\Theta}_s^{(\ell)}, \boldsymbol{\Theta}_t^{(\ell)} \in \mathbb{R}^{d \times d}$ are learnable transformation matrices, and $\boldsymbol{a}^{(\ell)} \in \mathbb{R}^d$ is a learnable projection vector.

\paragraph{Zone-Level Semantic Representation.}

To study the correlation between road segments that belong to the same functional zone, we first employ a multilevel graph partitioning algorithm~\cite{sanders2013think} to divide the road network into $k$ zones based on its topological structure, ensuring that each road segment belongs to a single traffic zone. Each traffic zone contains several road segments, and the number of road segments in different zones is relatively balanced. Then for a given traffic zone $z_i$, we assign an embedding vector $\boldsymbol{z}_i \in \mathbb{R}^d$ to its ID.

After defining the traffic zones, our goal is to capture the relationships between adjacent zones. Under the assumption that a higher traffic flow between two zones indicates a stronger connection, we first calculate the traffic flow between adjacent zones using training data to construct the matrix $\boldsymbol{F} \in \mathbb{R}^{k \times k}$, where $\boldsymbol{F}_{ij}$ represents the traffic flow between zones $i$ and $j$. Using this matrix, we apply GCN~\cite{kipf2017semi} to effectively obtain zone-level representations from their neighborhoods. Let $\boldsymbol{Z}^{(\ell)} = [\boldsymbol{z}_1^{(\ell)}, \boldsymbol{z}_2^{(\ell)}, \ldots, \boldsymbol{z}_k^{(\ell)}]^\top \in \mathbb{R}^{k \times d}$ denote the matrix of contextual representations of the traffic zones at the $\ell$-th layer, then the update process can be expressed as:
\begin{equation}
\boldsymbol{Z}^{(\ell + 1)} = \hat{\boldsymbol{D}}^{-1/2} \hat{\boldsymbol{F}} \hat{\boldsymbol{D}}^{-1/2} \boldsymbol{Z}^{(\ell)} \boldsymbol{\Theta}.
\end{equation}
Here, $\hat{\boldsymbol{F}} = \boldsymbol{F} / \max(\boldsymbol{F}) + \boldsymbol{I}$ denotes the 0-1 normalized matrix $\boldsymbol{F}$ with inserted self-loops, $\hat{\boldsymbol{D}}_{ii} = \sum_{j=1}^{k}\hat{\boldsymbol{F}}_{ij}$ is its diagonal degree matrix, and $\boldsymbol{\Theta} \in \mathbb{R}^{d \times d}$ is a trainable weight matrix used for the linear transformation.

\subsection{Multi-Granularity Trajectory Encoder}

Trajectory data contains rich semantic information, but effectively extracting it involves overcoming challenges at various granularities. At a fine granularity, it requires precise modeling of spatio-temporal points within the trajectory. At a coarse granularity, it necessitates capturing the dependencies between these spatio-temporal points. To address these challenges, we propose the Multi-Granularity Trajectory Encoder, which integrates both levels of modeling to fully capture the trajectory's semantic information.

\paragraph{Spatio-temporal Point Semantics.}

For the $i$-th spatio-temporal point $x_i = (r_i, t_i)$ in the trajectory, let $\text{zone}(r_i)$ be the zone index of $r_i$. In the modeling of spatial features, we utilize the road- and zone-level road network representations obtained from the previous Road Network Encoder, denoted as $\boldsymbol{v}_{r_i}$ and $\boldsymbol{z}_{\text{zone}(r_i)}$, respectively. Subsequently, we utilize a gating unit to fuse the representations at different levels to  obtain the spatial representation $\boldsymbol{x}_i^{\text{spatial}} \in \mathbb{R}^d$:
\begin{equation}
\resizebox{.875\columnwidth}{!}{$
\boldsymbol{x}_i^\text{spatial} = \boldsymbol{v}_{r_i} + \textrm{Sigmoid} \big(\textrm{MLP}(\boldsymbol{v}_{r_i} \ \Vert \ \boldsymbol{z}_{\text{zone}(r_i)})\big) \cdot \boldsymbol{z}_{\text{zone}(r_i)},
$}
\end{equation}
where $\textrm{MLP}$ converts a vector of length $2d$ into a scalar. To model temporal features, we employ the Fourier encoding strategy~\cite{xu2020inductive} to obtain the temporal representation $\boldsymbol{x}_i^{\text{temporal}} \in \mathbb{R}^d$ for the $i$-th spatio-temporal point:
\begin{equation}
\boldsymbol{x}_i^{\text{temporal}} = \sqrt{1/2d} \ \big[\cos(\omega_l t_i), \sin(\omega_l t_i)\big]_{l=1}^{d/2}.
\end{equation}
By concatenating the two aforementioned vectors, we obtain the representation of the $i$-th spatio-temporal point in the trajectory, denoted as $\boldsymbol{x}_i \in \mathbb{R}^{2d}$:
\begin{equation}
\boldsymbol{x}_i = \boldsymbol{x}_i^{\text{spatial}} \ \Vert \ \boldsymbol{x}_i^{\text{temporal}}.
\end{equation}

\paragraph{Trajectory Semantics.}

After obtaining the representations of all spatio-temporal points in the trajectory, we employ a Decoder-Only Transformer~\cite{radford2018improving} to extract the semantic information embedded within the trajectory. To more accurately capture the spatio-temporal relationships between these points, we introduce a relative position encoding technique~\cite{shaw2018self} based on spatio-temporal distances. Let $(\boldsymbol{x}_1, \boldsymbol{x}_2, \cdots, \boldsymbol{x}_n)$ be the representations of the input trajectory, we first encode the spatio-temporal interval between the input $\boldsymbol{x_i}$ and $\boldsymbol{x_j}$ as vectors $\boldsymbol{a}_{ij} \in \mathbb{R}^{2d/N_h}$:
\begin{equation}
\boldsymbol{a}_{ij} = d(r_i, r_j) \theta_d \ \Vert \ \Delta{t}(t_i, t_j) \theta_t.
\end{equation}
Here, $d(r_i, r_j)$ represents the distance between road segment $r_i$ and $r_j$, $\Delta{t}(t_i, t_j)$ represents the time interval between timestamp $t_i$ and $t_j$; $\theta_d, \theta_t \in \mathbb{R}^{d/N_h}$ are projection vectors and $N_h$ is the number of heads. The spatio-temporal relative encodings $\boldsymbol{a}_{ij}$ are built separately for the key and value (\Cref{eq:self_attention_with_relative_position} and~(\ref{eq:self_attention_with_relative_position_weight_coefficient})) computation of the Decoder-Only Transformer, then the operation of one head in the multi-head self-attention is:
\begin{equation}
\boldsymbol{\tau}_{1:i}^h = \sum_{j=1}^{i} \alpha_{ij} (\boldsymbol{x}_j \boldsymbol{\Theta}_v + \boldsymbol{a}_{ij}^{v}),
\label{eq:self_attention_with_relative_position}
\end{equation}
where the attention coefficients $\alpha_{ij}$ are computed as follows:
\begin{equation}
\alpha_{ij} = \textrm{Softmax} \big(\boldsymbol{x}_i \boldsymbol{\Theta}_q (\boldsymbol{x}_j \boldsymbol{\Theta}_k + \boldsymbol{a}_{ij}^{k})^\top / d_k\big).
\label{eq:self_attention_with_relative_position_weight_coefficient}
\end{equation}
Here, $\boldsymbol{\Theta}_q, \boldsymbol{\Theta}_k, \boldsymbol{\Theta}_v \in \mathbb{R}^{2d \times 2d/N_h}$ are the learnable matrices for query, key and value projections, respectively. The remainder of ours aligns with the structure of the Transformer Decoder. For the aforementioned input spatio-temporal points, we denote the output of the Decoder-Only Transformer as $(\boldsymbol{\tau}_{1:1}^{\text{out}}, \boldsymbol{\tau}_{1:2}^{\text{out}}, \cdots, \boldsymbol{\tau}_{1:n}^{\text{out}})$, where $\boldsymbol{\tau}_{1:i}^{\text{out}}$ corresponds to the semantics of the input trajectory $x_{1:i}$.

\subsection{Destination-Oriented Navigator}

Given that human movement frequently demonstrates clear intentionality and destination-oriented behavior, it is essential to integrate destination guidance within the modeling framework. To this end, we propose a novel Destination-Oriented Navigator, which predicts the next spatio-temporal point by effectively integrating partial trajectory features with destination guidance. Let the current partial trajectory be denoted as $x_{1:i} = [x_1, x_2, \ldots, x_i]$. Additionally, let $R(r_i)$ represent the set of road segments that are reachable from the current road segment $r_i$. When predicting the probability of a candidate road segment $r_\text{c} \in R(r_i)$ as the next step, we consider not only the semantics of the current partial trajectory $\boldsymbol{\tau}_{1:i}^{\text{out}}$ and the representations of the candidate road segment $\boldsymbol{v}_{r_\text{c}}$, but also the feature of the destination zone $\boldsymbol{z}_{z_\text{dest}}$ and the metric characteristics from the candidate road segment to the destination $\boldsymbol{h}_{r_\text{c}, r_\text{dest}}$ (including distances and angles, more details in Appendix A.1).

We then utilize an additive attention mechanism~\cite{bahdanau2015neural} to integrate the aforementioned features. Specifically, the semantics of the partial trajectory $\boldsymbol{\tau}_{1:i}^{\text{out}} \in \mathbb{R}^{2d}$ and the representation of the destination zone $\boldsymbol{z}_{z_\text{dest}} \in \mathbb{R}^{d}$ are used as queries, while the representations of candidate road segments $\boldsymbol{v}_{r_\text{c}} \in \mathbb{R}^d$ and the metric information from the candidate road segment to the destination $\boldsymbol{h}_{r_\text{c}, r_\text{dest}} \in \mathbb{R}^{2d}$ are used as keys, then the logit of the candidate road segment $r_\text{c}$ can be written as:
\begin{equation}
\resizebox{.875\columnwidth}{!}{$
p_{r_\text{c}} = \tanh \big((\boldsymbol{\tau}_{1:i}^{\text{out}} \ \Vert \ \boldsymbol{z}_{\text{z}_{\text{dest}}}) \mathbf{W}_q + (\boldsymbol{v}_{r_\text{c}} \ \Vert \ \boldsymbol{h}_{r_\text{c}, r_\text{dest}}) \mathbf{W}_k\big) \mathbf{w}_{v}^{\top},
$}
\end{equation}
where $\mathbf{W}_q \in \mathbb{R}^{3d \times d}, \mathbf{W}_k \in \mathbb{R}^{3d \times d}, \mathbf{w}_v \in \mathbb{R}^{d}$ are the learnable parameters for query, key, and value, respectively. After applying the Softmax, the probability can be obtained:
\begin{equation}
\hat{P_\theta}(r_\text{c} | x_{1:i}, r_\text{dest}) = \frac{\exp (p_{r_\text{c}})}{\sum_{r_\text{c}^\prime \in R(r_i)} \exp(p_{r_\text{c}})}.
\end{equation}

To predict the timestamp $t_{i+1}$ for the aforementioned candidate road segment $r_\text{c}$, we reformulate it as predicting the time interval to the next position $\Delta t_{i+1} = t_{i+1} - t_i$. This prediction utilizes both the semantics of the partial trajectory $x_{1:i}$ and the features of the candidate road segment $r_\text{c}$, employing a MLP to yield a single numerical output:
\begin{equation}
\Delta \hat{t}_{i+1} = \textrm{MLP}(\boldsymbol{\tau}_{1:i}^{\text{out}} \ \Vert \ \boldsymbol{v}_{r_\text{c}}).
\end{equation}

\subsection{End-to-End Learning}

\paragraph{Optimization.}

During training, we predict the next reachable road segment and the corresponding time interval, based on partial real trajectories $x_{1:i}$ and the destination $r_\text{dest}$. The negative log-likelihood loss $\mathcal{L}_r$ is used for road segment prediction, while the mean absolute error loss $\mathcal{L}_t$ is used for interval time prediction. We add them together to optimize the model, written as:
\begin{equation}
\resizebox{.875\columnwidth}{!}{$
\mathcal{L} = \frac{1}{n-1} \displaystyle\sum_{i=1}^{n-1} \underbrace{-\log \hat{P_\theta}(r_{i+1} | x_{1:i}, r_\text{dest})}_{\mathcal{L}_r} + \underbrace{\lvert \Delta \hat{t}_{i+1} - \Delta t_{i+1} \rvert}_{\mathcal{L}_t}.
$}
\end{equation}

\paragraph{Generation.}

Given the conditional information $(r_\text{org}, t_\text{org}, r_\text{dest})$, we search the trajectory with the maximum probability as the final generated trajectory, as described in \Cref{eq:trajectory_with_the_maximum_probability}. In practice, a heap is utilized to accelerate the process (more details in Appendix A.2).

\section{Experiments}

\begin{table*}[t]
\centering
\small
\setlength{\tabcolsep}{1mm}
\begin{tabular}{lcccccccccccc}
\toprule
& \multicolumn{4}{c|}{Beijing} & \multicolumn{4}{c|}{Porto} & \multicolumn{4}{c}{San Francisco}\\
\cmidrule(lr){2-13}
\multirow{2.5}{*}{Methods} & \multicolumn{3}{c|}{Global ($\mathrel{\downarrow}$)} & \multicolumn{1}{c|}{Local ($\mathrel{\downarrow}$)} & \multicolumn{3}{c|}{Global ($\mathrel{\downarrow}$)} & \multicolumn{1}{c|}{Local ($\mathrel{\downarrow}$)} & \multicolumn{3}{c|}{Global ($\mathrel{\downarrow}$)} & Local ($\mathrel{\downarrow}$)\\
\cmidrule(lr){2-13}
& Distance & Radius & \multicolumn{1}{c|}{Duration} & \multicolumn{1}{c|}{Hausdorff} & Distance & Radius & \multicolumn{1}{c|}{Duration} & \multicolumn{1}{c|}{Hausdorff} & Distance & Radius & \multicolumn{1}{c|}{Duration} & Hausdorff\\
\midrule
Markov & 0.0048 & 0.0168 & \ding{55} & \multicolumn{1}{c|}{0.8164} & 0.0047 & 0.0294 & \ding{55} & \multicolumn{1}{c|}{0.7158} & 0.0052 & 0.0250 & \ding{55} & 0.7546\\
Dijkstra & 0.0062 & 0.0064 & \ding{55} & \multicolumn{1}{c|}{0.6239} & 0.0177 & 0.0099 & \ding{55} & \multicolumn{1}{c|}{\underline{0.6011}} & 0.0128 & 0.0060 & \ding{55} & \underline{0.5567}\\
SeqGAN & 0.0068 & 0.0077 & \ding{55} & \multicolumn{1}{c|}{0.6982} & 0.0089 & 0.0082 & \ding{55} & \multicolumn{1}{c|}{0.7049} & 0.0092 & 0.0043 & \ding{55} & 0.6959\\
SVAE & 0.0077 & 0.0124 & \ding{55} & \multicolumn{1}{c|}{0.7180} & 0.0095 & 0.0250 & \ding{55} & \multicolumn{1}{c|}{0.6669} & 0.0188 & 0.0422 & \ding{55} & 0.5908\\
MoveSim & 0.3169 & 0.2091 & \ding{55} & \multicolumn{1}{c|}{4.3434} & 0.0929 & 0.1015 & \ding{55} & \multicolumn{1}{c|}{1.3911} & 0.1464 & 0.0946 & \ding{55} & 1.5704\\
TSG & 0.4498 & 0.1471 & \ding{55} & \multicolumn{1}{c|}{0.8636} & 0.1769 & 0.3037 & \ding{55} & \multicolumn{1}{c|}{0.5676} & 0.3464 & 0.0952 & \ding{55} & 0.8720\\
TrajGen & 0.2750 & 0.1553 & \ding{55} & \multicolumn{1}{c|}{3.5120} & 0.2305 & 0.2287 & \ding{55} & \multicolumn{1}{c|}{1.3774} & 0.2895 & 0.0652 & \ding{55} & 1.7050\\
DiffTraj & 0.0033 & 0.0078 & \ding{55} & \multicolumn{1}{c|}{0.6483} & 0.0070 & 0.0066 & \ding{55} & \multicolumn{1}{c|}{0.6005} & \underline{0.0040} & 0.0384 & \ding{55} & 0.6196\\
STEGA & 0.0090 & 0.0331 & 0.2858 & \multicolumn{1}{c|}{0.7473} & 0.0128 & 0.0877 & 0.2239 & \multicolumn{1}{c|}{0.6231} & 0.0155 & 0.1376 & 0.3468 & 0.5984\\
TS-TrajGen & 0.0172 & 0.0059 & \underline{0.2580} & \multicolumn{1}{c|}{0.9618} & 0.0050 & 0.0052 & 0.2023 & \multicolumn{1}{c|}{0.7153} & 0.0143 & 0.0062 & \underline{0.2931} & 0.7605\\
\cmidrule(lr){1-13}
Markov* & 0.0034 & 0.0037 & \ding{55} & \multicolumn{1}{c|}{\underline{0.6086}} & \textbf{0.0041} & 0.0092 & \ding{55} & \multicolumn{1}{c|}{0.6410} & 0.0049 & \underline{0.0037} & \ding{55} & 0.6161\\
SeqGAN* & \underline{0.0029} & \underline{0.0032} & \ding{55} & \multicolumn{1}{c|}{0.6099} & 0.0055 & \underline{0.0039} & \ding{55} & \multicolumn{1}{c|}{0.6903} & 0.0055 & 0.0043 & \ding{55} & 0.6605\\
MoveSim* & 0.0651 & 0.0099 & \ding{55} & \multicolumn{1}{c|}{1.3311} & 0.0309 & 0.0108 & \ding{55} & \multicolumn{1}{c|}{0.9110} & 0.0292 & 0.0074 & \ding{55} & 0.7708\\
STEGA* & 0.0086 & 0.0054 & 0.2747 & \multicolumn{1}{c|}{0.6923} & 0.0528 & 0.0353 & \underline{0.1819} & \multicolumn{1}{c|}{1.0897} & 0.0112 & 0.0082 & 0.3820 & 0.7216\\
\cmidrule(lr){1-13}
Ours w/o RNE & 0.0024 & 0.0026 & 0.0274 & \multicolumn{1}{c|}{0.5694} & 0.0050 & 0.0038 & 0.0242 & \multicolumn{1}{c|}{0.5993} & 0.0037 & 0.0036 & 0.0499 & 0.5403\\
Ours w/o TrajE & 0.0027 & 0.0029 & 0.0268 & \multicolumn{1}{c|}{0.5650} & 0.0051 & 0.0041 & 0.0219 & \multicolumn{1}{c|}{0.5956} & 0.0040 & 0.0039 & 0.0487 & 0.5381\\
Ours w/o Nav & 0.0029 & 0.0030 & 0.0259 & \multicolumn{1}{c|}{0.5704} & 0.0055 & 0.0043 & 0.0237 & \multicolumn{1}{c|}{0.6263} & 0.0045 & 0.0040 & 0.0359 & 0.5661\\
Ours & \textbf{0.0024} & \textbf{0.0025} & \textbf{0.0245} & \multicolumn{1}{c|}{\textbf{0.5503}} & \underline{0.0045} & \textbf{0.0033} & \textbf{0.0197} & \multicolumn{1}{c|}{\textbf{0.5746}} & \textbf{0.0033} & \textbf{0.0034} & \textbf{0.0249} & \textbf{0.5351}\\
\bottomrule
\end{tabular}

\caption{Average performance of 5 random seeds (0 to 4) on three real-world trajectory datasets in terms of global and local level metrics. The method names followed by an asterisk (*) indicate the corresponding search versions. The best one is denoted by \textbf{boldface} and the second-best is denoted by \underline{underline}. Unsupported metrics are denoted by \ding{55}. $\mathrel{\downarrow}$ denotes lower is better.}
\label{tab:performance_comparison}
\end{table*}

We conducted extensive experiments on three real-world trajectory datasets to validate the performance of HOSER. This section outlines the basic experimental setup and the main experimental results, while additional details are available in the Appendix due to space constraints. All experiments are conducted on a single NVIDIA RTX A6000 GPU.

\subsection{Experimental Setups}

\paragraph{Datasets.}

We assess the performance of HOSER and other baselines using three trajectory datasets from Beijing, Porto, and San Francisco. Each dataset is randomly split into training, validation, and test sets in a 7:1:2 ratio. Further dataset details are provided in Appendix B.1.

\paragraph{Evaluation Metrics.}

To comprehensively evaluate the quality of synthetic trajectories, we compare the trajectories generated by HOSER and other baselines with real trajectories from the following global and local perspectives, which follow the design in \cite{jiang2023continuous, wang2024stega}. For more details, please refer to Appendix B.2.

From the global perspective, we measure the overall distribution of the trajectories using the following three metrics: \textit{Distance}, \textit{Radius}, and \textit{Duration}. To obtain quantitative results, we employ Jensen-Shannon divergence (JSD) to measure the distribution similarity of the three metrics.

From the local perspective, we exclusively compare the similarity between real and generated trajectories that have the same OD pairs, using the following three metrics for evaluation, i.e., \textit{Hausdorff distance}, \textit{DTW} and \textit{EDR}.

\paragraph{Baselines.}

We compare HOSER with a series of baselines, including both traditional methods and a suite of deep learning-based methods. The former includes Markov~\cite{gambs2012next} and Dijkstra's algorithm~\cite{dijkstra1959note}, while the latter comprises SeqGAN~\cite{yu2017seqgan}, SVAE~\cite{huang2019variational}, MoveSim~\cite{feng2020learning}, TSG~\cite{wang2021large}, TrajGen~\cite{cao2021generating}, DiffTraj~\cite{zhu2023difftraj}, STEGA~\cite{wang2024stega}, and TS-TrajGen~\cite{jiang2023continuous}. See Appendix B.3 for more details.

\subsection{Overall Performance}

\paragraph{Quantitative Analysis.}

The global and local metrics on three real-world trajectory datasets are shown in \Cref{tab:performance_comparison}. Due to space limitations, the DTW and EDR metrics for these three datasets are provided in Appendix C.1. The results demonstrate that compared to other \emph{state-of-the-art} baselines, the trajectories generated by HOSER are closer to real-world trajectories in terms of both global and local similarity. This satisfactory result can be largely attributed to our comprehensive modeling of human mobility patterns. Among the baseline methods, DiffTraj demonstrates superior performance due to its advanced diffusion architecture. TS-TrajGen also achieves commendable results by integrating neural networks with the A* algorithm to model human mobility patterns. Additionally, and somewhat unexpectedly, Dijkstra's algorithm outperforms most deep learning-based approaches. This can be explained by the fact that people typically choose the quickest route to their destination based on their personal experience, and this route often approximates the shortest route~\cite{yuan2010tdrive}. However, due to factors such as traffic signals and road congestion, the quickest route does not always align with the shortest. HOSER effectively accounts for these discrepancies through its novel network architecture, resulting in superior performance and further highlighting the significance of modeling human mobility patterns holistically.

\paragraph{Discussion on Baselines with the Search Algorithm.}

Since our method, along with TS-TrajGen, utilizes a search-based paradigm (as described in \Cref{eq:trajectory_with_the_maximum_probability}) to find the optimal trajectory with the highest probability between OD pairs, rather than relying solely on autoregressively generating entire trajectories based on previously generated points, we conduct additional experiments by modifying some baseline models to a search-based paradigm to investigate the impact of this paradigm on the effectiveness of trajectory generation. Specifically, we reformulate Markov, SeqGAN, MoveSim, and STEGA into search-based forms, appending ``*'' to denote the corresponding variants. It can be observed from \Cref{tab:performance_comparison} that after switching to a search-based paradigm, their performance has improved to some extent compared to the original approach. However, since they do not comprehensively model the semantics of human movement and instead simply use the partially generated trajectory to predict the next spatio-temporal point, there remains a gap between their performance and ours. This also indirectly suggests that the effectiveness of our method is not solely due to the adoption of the search-based paradigm.

\begin{figure}[t]
\centering
\includegraphics[width=1.0\columnwidth]{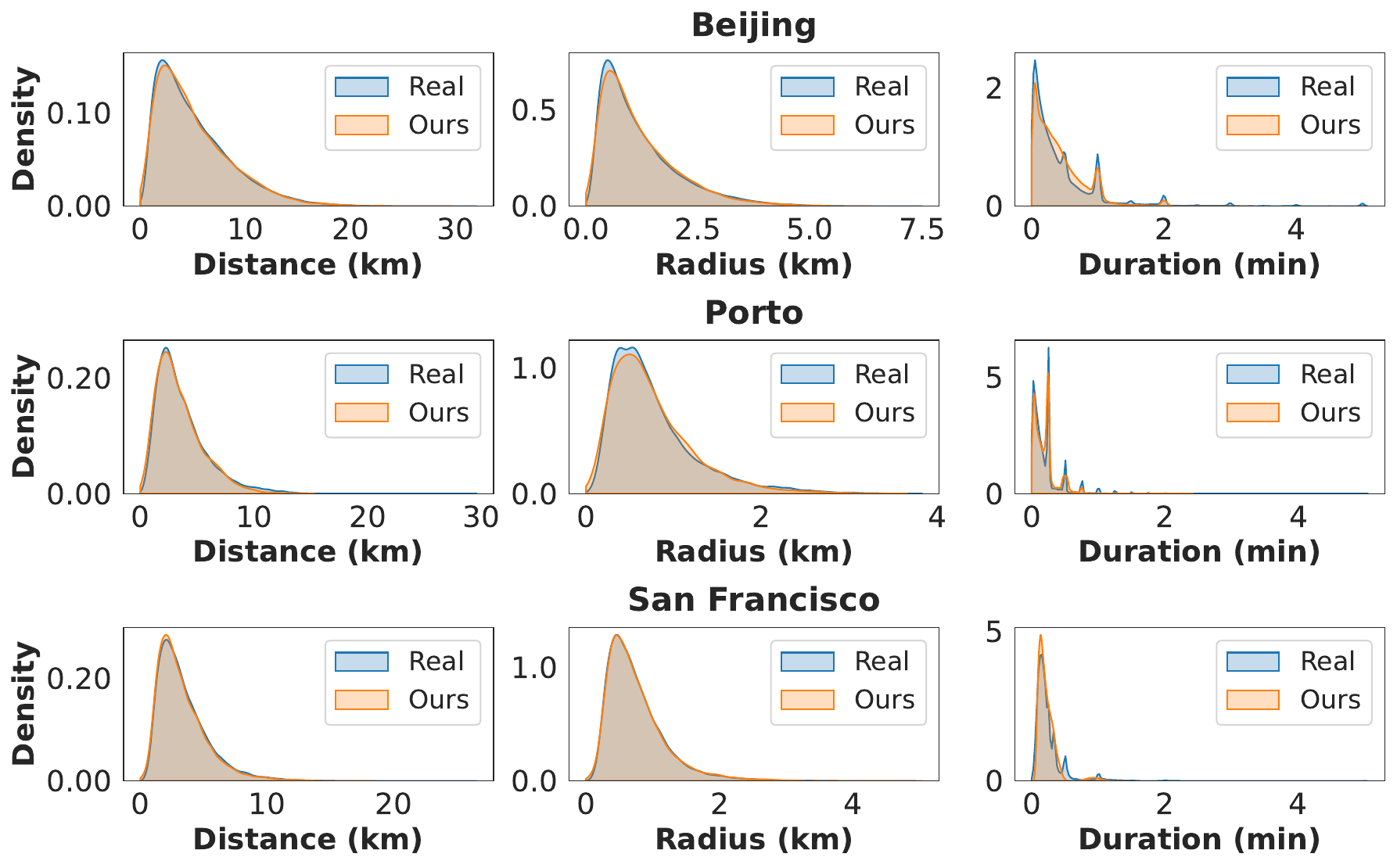}
\caption{Visualization of metrics distributions.}
\label{fig:vis_distro}
\end{figure}

\paragraph{Ablation Studies.}

HOSER comprises three key modules: the Road Network Encoder which models the road network at different levels, 
the Trajectory Encoder which extracts the semantics of partial trajectories, and the Navigator which seamlessly integrates destination guidance. To assess the contribution of each module to the overall performance, we perform ablation studies on three HOSER variants, each corresponding to the removal of one module, denoted as Ours w/o RNE, Ours w/o TrajE, and Ours w/o Nav, respectively (please refer to Appendix C.1 for more details).

\Cref{tab:performance_comparison} shows that performance declines when any module is removed, indicating their necessity for high-fidelity trajectory generation. Among them, the removal of the Navigator has the most significant impact on model performance, underscoring the importance of incorporating destination guidance in trajectory generation. Moreover, the significant drop in the Duration metric after removing the Road Network Encoder highlights the critical role of road network representation in accurately predicting travel time. Lastly, the removal of the Trajectory Encoder results in a decline across all performance metrics, indicating that generating reliable trajectories requires not only destination information but also historical trajectory data.

\paragraph{Visualization Analysis.}

To intuitively compare the similarity between real and generated trajectories, we visualize the distribution of metrics including \textit{Distance}, \textit{Radius} and \textit{Duration} of the trajectories, as shown in \Cref{fig:vis_distro}. Specifically, for the \textit{Distance} and \textit{Radius} metrics, the generated data not only captures the peak values but also aligns well with the long-tail distributions of the real data. For the \textit{Duration} metric, the synthetic data successfully replicates the multimodal characteristics observed in the real data, further confirming the reliability of the synthetic data.

\begin{figure}[t]
\centering
\includegraphics[width=1.0\columnwidth]{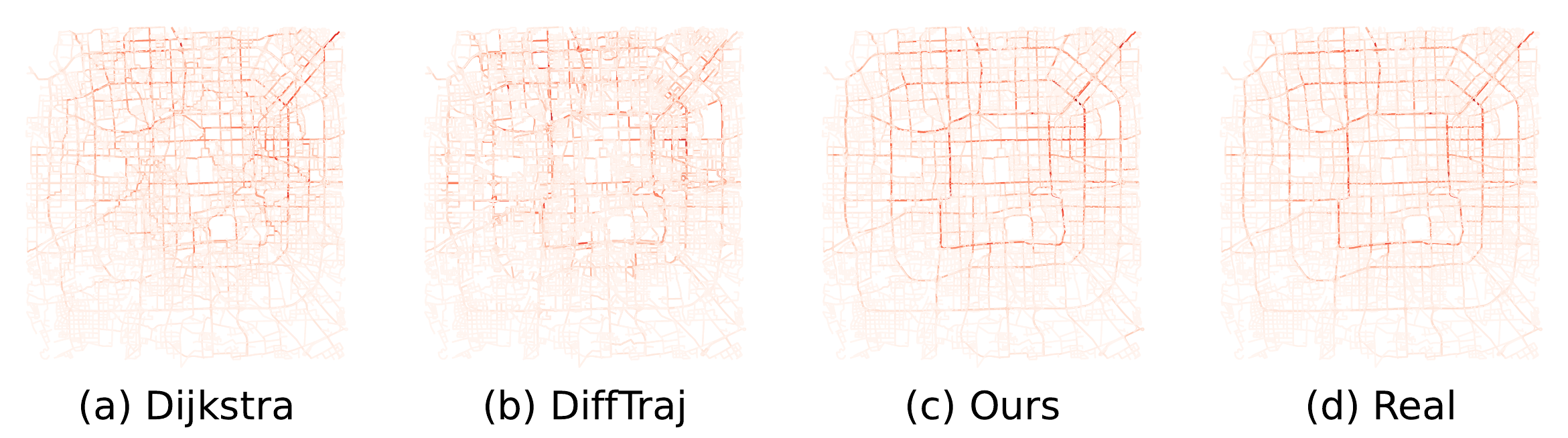}
\caption{Visualization of the trajectories in Beijing (a larger view for Beijing, as well as for the other two cities, can be found in Appendix C.1).}
\label{fig:vis_heatmap_beijing}
\end{figure}

We also visualize both the real trajectories and the generated trajectories to facilitate a more intuitive comparison. \Cref{fig:vis_heatmap_beijing} presents a heatmap illustrating the distribution of real trajectories alongside those generated by the top three methods in Beijing. Since Dijkstra's algorithm directly uses the shortest path between OD pairs to generate trajectories, the frequency of road segment access is relatively uniform. In addition, DiffTraj fails to fully consider the topological structure of the road network, resulting in a significant discrepancy from actual data. In contrast, our method nearly matches the original trajectories perfectly, indicating a marked improvement over other methods.

\subsection{Utility of Generated Data}

Since the generated trajectories are ultimately used to analyze human mobility patterns, their utility determines whether the data generation method is feasible. Here, we evaluate the utility of the generated trajectories through a well-known location prediction task. We train three advanced prediction models: DeepMove~\cite{feng2018deepmove}, Flashback~\cite{yang2020location}, and LSTPM~\cite{sun2020go}, using both real and generated data, and compare their performance. As shown in \Cref{tab:data_utility_comparison}, DeepMove and LSTPM perform comparably with synthetic and real data, while FlashBack shows slight deviations due to its reliance on timestamp information, indicating room for improvement. Nevertheless, these results highlight the potential of generated trajectories as viable substitutes for real data (please refer to Appendix C.2 for the results of other baselines).

\begin{table}[t]
\centering
\small
\begin{tabular}{llcc}
\toprule
Datasets & Methods & Acc@5 & MRR\\
\midrule
\multirow{3}{*}{Beijing} & DeepMove & 0.776 / 0.804 & 0.697 / 0.728\\
 & Flashback & 0.749 / 0.782 & 0.676 / 0.706\\
 & LSTPM & 0.761 / 0.795 & 0.694 / 0.713\\
\cmidrule(lr){1-4}
\multirow{3}{*}{Porto} & DeepMove & 0.888 / 0.929 & 0.758 / 0.780\\
 & Flashback & 0.812 / 0.895 & 0.698 / 0.761\\
 & LSTPM & 0.860 / 0.914 & 0.741 / 0.778\\
\cmidrule(lr){1-4}
\multirow{3}{*}{San Francisco} & DeepMove & 0.797 / 0.847 & 0.673 / 0.698\\
 & Flashback & 0.746 / 0.815 & 0.625 / 0.685\\
 & LSTPM & 0.774 / 0.816 & 0.667 / 0.680\\
\bottomrule
\end{tabular}
\caption{Comparison of data utility based on location prediction task, results are expressed as (generated / real).}
\label{tab:data_utility_comparison}
\end{table}

\subsection{Few-Shot and Zero-Shot Learning Tests}

Considering the scarcity of real-world trajectory data, the few-shot and zero-shot capabilities of the trajectory generation model are crucial. Therefore, we evaluate the few-shot and zero-shot capabilities of HOSER.

\begin{figure}[t]
    \centering
    \includegraphics[width=\columnwidth]{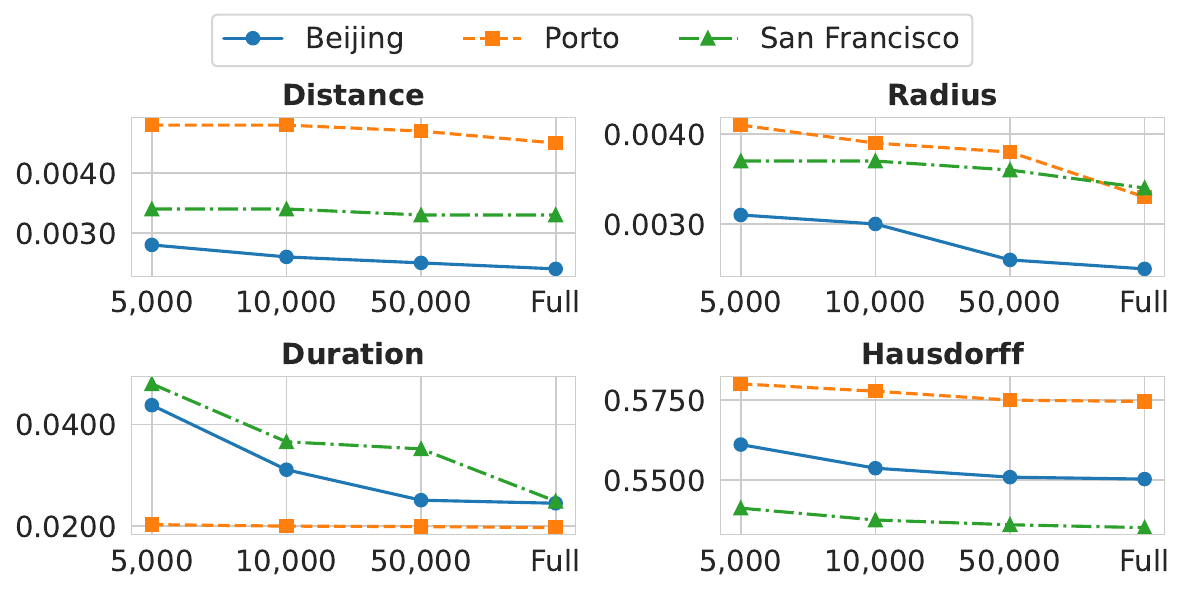}
    \caption{HOSER's performance with varying amounts of training data across three trajectory datasets. ``Full'' denotes the complete dataset, with sizes of \num{629380}, \num{481359}, and \num{205116} for Beijing, Porto, and San Francisco, respectively.}
    \label{fig:few_shot_perf}
\end{figure}

For few-shot learning, we randomly sample \num{5000}, \num{10000}, and \num{50000} trajectories for training and compare the performance of the generated trajectories. As shown in \Cref{fig:few_shot_perf}, our model’s precise representation of road networks and incorporation of human mobility patterns enables strong performance even with limited data, which improves as the size of the training dataset increases.

For zero-shot learning, among the baselines, Dijkstra, TS-TrajGen, and DiffTraj perform well in general trajectory generation tasks. However, as TS-TrajGen lacks support for zero-shot learning, we compare HOSER specifically with Dijkstra and DiffTraj in this context. As shown in \Cref{tab:transferability_vis}, HOSER excels in zero-shot learning tasks due to its holistic semantic modeling of human mobility patterns, which effectively captures and leverages the universality of policies employed in human mobility, enhancing its generalizability.

\begin{table}[t]
\centering
\small
\begin{tabular}{lcccc}
\toprule
\multirow{2.5}{*}{Methods}& \multicolumn{3}{c|}{Global ($\mathrel{\downarrow}$)} & \multicolumn{1}{c}{Local ($\mathrel{\downarrow}$)}\\
\cmidrule(lr){2-5}
 & Distance & Radius & \multicolumn{1}{c|}{Duration} & Hausdorff\\
\midrule
Dijkstra & \underline{0.0177} & \underline{0.0099} & \ding{55} & \underline{0.6011}\\
DiffTraj & 0.0633 & 0.2521 & \ding{55} & 0.8023\\
HOSER & \textbf{0.0052} & \textbf{0.0053} & \textbf{0.0223} & \textbf{0.5843}\\
\bottomrule
\end{tabular}
\caption{Results of zero-shot learning. DiffTraj and HOSER are trained in Beijing and generated in Porto, while Dijkstra is generated directly in Porto.}
\label{tab:transferability_vis}
\end{table}

\section{Related Work}

\paragraph{Trajectory Generation.}

Trajectory synthesis methods fall into two categories: model-based and model-free. Model-based methods~\cite{song2010modelling, jiang2016timegeo} assume interpretable mobility patterns but often oversimplify real-world complexity. Model-free methods are further classified into grid-based, coordinate point-based, and road segment-based approaches. Grid-based methods generate matrix trajectory data by dividing the map into grids~\cite{ouyang2018non, cao2021generating}. Coordinate point-based methods map GPS points to high-dimensional spaces via linear transformations and apply generative models~\cite{kingma2014auto, goodfellow2014generative, ho2020denoising, liu2024interaction}, including VAE~\cite{huang2019variational}, GAN~\cite{wang2021large}, and diffusion-based models~\cite{zhu2023difftraj, zhu2023synmob, zhu2024controltraj}. Road segment-based methods~\cite{feng2020learning, cao2021generating, jiang2023continuous, wang2024stega} embed road segments as tokens. However, existing methods struggle to balance different aspects of human mobility patterns.

\paragraph{Road Network Representation Learning.}

Road networks are crucial for intelligent transportation tasks like spatial query processing~\cite{huang2021learning, zhao2022rne, chang2023contrastive}, travel time estimation~\cite{yuan2022route}, and traffic forecasting~\cite{guo2021learning}. Early studies~\cite{jepsen2018network, jepsen2019graph, wang2019learning, wang2020representation, wu2020learning} leverage GNNs~\cite{kipf2017semi, velivckovic2018graph, zheng2022transition, zheng2023temporal} for road network representation learning. Recent work~\cite{chen2021robust, mao2022jointly, schestakov2023road, zhang2023road, chen2024semantic} enhances road representations by integrating trajectory data. Nonetheless, applying these methods to trajectory generation remains challenging, demanding specialized integration models.

\section{Conclusion}

This paper introduces HOSER, a novel trajectory generation framework enhanced with holistic semantic representation, which incorporates multi-level road network encoding, multi-granularity trajectory representation, and destination guidance modeling. Extensive experiments demonstrate that our method surpasses \emph{state-of-the-art} baselines in global and local similarity metrics. The synthetic trajectories are effective for downstream tasks, demonstrating their potential as real-data substitutes. Additionally, HOSER performs well in few-shot and zero-shot learning. In the future, we will investigate the division of dense spatio-temporal points along a trajectory into coarse-grained activity sequences and fine-grained road segment sequences, facilitating the semantic representations of trajectories at varying scales.

\FloatBarrier

\section*{Acknowledgments}

This work is supported by the Zhejiang Province ``JianBingLingYan+X'' Research and Development Plan (2024C01114), Zhejiang Province High-Level Talents Special Support Program ``Leading Talent of Technological Innovation of Ten-Thousands Talents Program'' (No.2022R52046), the Fundamental Research Funds for the Central Universities (No.226-2024-00058), and the Scientific Research Fund of Zhejiang Provincial Education Department (Grant No.Y202457035). Also, we thank Bayou Tech (Hong Kong) Limited for providing the data used in this paper free of charge.

\bibliography{aaai25}

\begin{thebibliography}{64}
\providecommand{\natexlab}[1]{#1}

\bibitem[{Bahdanau, Cho, and Bengio(2015)}]{bahdanau2015neural}
Bahdanau, D.; Cho, K.; and Bengio, Y. 2015.
\newblock Neural machine translation by jointly learning to align and
  translate.
\newblock In \emph{{ICLR}}.

\bibitem[{Bao et~al.(2017)Bao, He, Ruan, Li, and Zheng}]{bao2017planning}
Bao, J.; He, T.; Ruan, S.; Li, Y.; and Zheng, Y. 2017.
\newblock Planning bike lanes based on sharing-bikes' trajectories.
\newblock In \emph{{SIGKDD}}.

\bibitem[{Brody, Alon, and Yahav(2022)}]{brody2022attentive}
Brody, S.; Alon, U.; and Yahav, E. 2022.
\newblock How attentive are graph attention networks?
\newblock In \emph{{ICLR}}.

\bibitem[{Cao and Li(2021)}]{cao2021generating}
Cao, C.; and Li, M. 2021.
\newblock Generating mobility trajectories with retained data utility.
\newblock In \emph{{SIGKDD}}.

\bibitem[{Chang et~al.(2023)Chang, Qi, Liang, and Tanin}]{chang2023contrastive}
Chang, Y.; Qi, J.; Liang, Y.; and Tanin, E. 2023.
\newblock Contrastive trajectory similarity learning with dual-feature
  attention.
\newblock In \emph{{ICDE}}.

\bibitem[{Chen, {\"O}zsu, and Oria(2005)}]{chen2005robust}
Chen, L.; {\"O}zsu, M.~T.; and Oria, V. 2005.
\newblock Robust and fast similarity search for moving object trajectories.
\newblock In \emph{{SIGMOD}}.

\bibitem[{Chen et~al.(2024{\natexlab{a}})Chen, Liang, Zhu, Chang, Luo, Wen, Li,
  Yu, Wen, Chen et~al.}]{chen2024deep}
Chen, W.; Liang, Y.; Zhu, Y.; Chang, Y.; Luo, K.; Wen, H.; Li, L.; Yu, Y.; Wen,
  Q.; Chen, C.; et~al. 2024{\natexlab{a}}.
\newblock Deep learning for trajectory data management and mining: A survey and
  beyond.
\newblock \emph{arXiv preprint arXiv:2403.14151}.

\bibitem[{Chen et~al.(2024{\natexlab{b}})Chen, Li, Cong, Bao, and
  Long}]{chen2024semantic}
Chen, Y.; Li, X.; Cong, G.; Bao, Z.; and Long, C. 2024{\natexlab{b}}.
\newblock Semantic-enhanced representation learning for road networks with
  temporal dynamics.
\newblock \emph{arXiv preprint arXiv:2403.11495}.

\bibitem[{Chen et~al.(2021)Chen, Li, Cong, Bao, Long, Liu, Chandran, and
  Ellison}]{chen2021robust}
Chen, Y.; Li, X.; Cong, G.; Bao, Z.; Long, C.; Liu, Y.; Chandran, A.~K.; and
  Ellison, R. 2021.
\newblock Robust road network representation learning: When traffic patterns
  meet traveling semantics.
\newblock In \emph{{CIKM}}.

\bibitem[{Dijkstra(1959)}]{dijkstra1959note}
Dijkstra, E.~W. 1959.
\newblock A note on two problems in connexion with graphs.
\newblock \emph{{Numer. Math.}}, 1: 269--271.

\bibitem[{Feng et~al.(2018)Feng, Li, Zhang, Sun, Meng, Guo, and
  Jin}]{feng2018deepmove}
Feng, J.; Li, Y.; Zhang, C.; Sun, F.; Meng, F.; Guo, A.; and Jin, D. 2018.
\newblock Deepmove: Predicting human mobility with attentional recurrent
  networks.
\newblock In \emph{{WWW}}.

\bibitem[{Feng et~al.(2020)Feng, Yang, Xu, Yu, Wang, and Li}]{feng2020learning}
Feng, J.; Yang, Z.; Xu, F.; Yu, H.; Wang, M.; and Li, Y. 2020.
\newblock Learning to simulate human mobility.
\newblock In \emph{{SIGKDD}}.

\bibitem[{Gambs, Killijian, and del Prado~Cortez(2012)}]{gambs2012next}
Gambs, S.; Killijian, M.-O.; and del Prado~Cortez, M.~N. 2012.
\newblock Next place prediction using mobility Markov chains.
\newblock In \emph{Proceedings of the first workshop on measurement, privacy,
  and mobility}.

\bibitem[{Gonzalez, Hidalgo, and Barabasi(2008)}]{gonzalez2008understanding}
Gonzalez, M.~C.; Hidalgo, C.~A.; and Barabasi, A.-L. 2008.
\newblock Understanding individual human mobility patterns.
\newblock \emph{{Nature}}, 453(7196): 779--782.

\bibitem[{Goodfellow et~al.(2014)Goodfellow, Pouget-Abadie, Mirza, Xu,
  Warde-Farley, Ozair, Courville, and Bengio}]{goodfellow2014generative}
Goodfellow, I.; Pouget-Abadie, J.; Mirza, M.; Xu, B.; Warde-Farley, D.; Ozair,
  S.; Courville, A.; and Bengio, Y. 2014.
\newblock Generative adversarial nets.
\newblock In \emph{{NeurIPS}}.

\bibitem[{Guo et~al.(2021)Guo, Lin, Wan, Li, and Cong}]{guo2021learning}
Guo, S.; Lin, Y.; Wan, H.; Li, X.; and Cong, G. 2021.
\newblock Learning dynamics and heterogeneity of spatial-temporal graph data
  for traffic forecasting.
\newblock \emph{{IEEE Trans. Knowl. Data Eng.}}, 34(11): 5415--5428.

\bibitem[{Han et~al.(2021)Han, Wang, Yao, Shang, and Zhang}]{han2021graph}
Han, P.; Wang, J.; Yao, D.; Shang, S.; and Zhang, X. 2021.
\newblock A graph-based approach for trajectory similarity computation in
  spatial networks.
\newblock In \emph{{SIGKDD}}.

\bibitem[{Ho, Jain, and Abbeel(2020)}]{ho2020denoising}
Ho, J.; Jain, A.; and Abbeel, P. 2020.
\newblock Denoising diffusion probabilistic models.
\newblock In \emph{{NeurIPS}}.

\bibitem[{Huang et~al.(2019)Huang, Song, Fan, Jiang, Shibasaki, Zhang, Wang,
  and Kato}]{huang2019variational}
Huang, D.; Song, X.; Fan, Z.; Jiang, R.; Shibasaki, R.; Zhang, Y.; Wang, H.;
  and Kato, Y. 2019.
\newblock A variational autoencoder based generative model of urban human
  mobility.
\newblock In \emph{{MIPR}}.

\bibitem[{Huang et~al.(2021)Huang, Wang, Zhao, and Li}]{huang2021learning}
Huang, S.; Wang, Y.; Zhao, T.; and Li, G. 2021.
\newblock A learning-based method for computing shortest path distances on road
  networks.
\newblock In \emph{{ICDE}}.

\bibitem[{Jepsen, Jensen, and Nielsen(2019)}]{jepsen2019graph}
Jepsen, T.~S.; Jensen, C.~S.; and Nielsen, T.~D. 2019.
\newblock Graph convolutional networks for road networks.
\newblock In \emph{{SIGSPATIAL}}.

\bibitem[{Jepsen et~al.(2018)Jepsen, Jensen, Nielsen, and
  Torp}]{jepsen2018network}
Jepsen, T.~S.; Jensen, C.~S.; Nielsen, T.~D.; and Torp, K. 2018.
\newblock On network embedding for machine learning on road networks: A case
  study on the danish road network.
\newblock In \emph{Big Data}.

\bibitem[{Jiang et~al.(2016)Jiang, Yang, Gupta, Veneziano, Athavale, and
  Gonz{\'a}lez}]{jiang2016timegeo}
Jiang, S.; Yang, Y.; Gupta, S.; Veneziano, D.; Athavale, S.; and Gonz{\'a}lez,
  M.~C. 2016.
\newblock The TimeGeo modeling framework for urban mobility without travel
  surveys.
\newblock \emph{{Proc. Natl. Acad. Sci. U.S.A.}}, 113(37): E5370--E5378.

\bibitem[{Jiang et~al.(2023)Jiang, Zhao, Wang, and Jiang}]{jiang2023continuous}
Jiang, W.; Zhao, W.~X.; Wang, J.; and Jiang, J. 2023.
\newblock Continuous trajectory generation based on two-stage GAN.
\newblock In \emph{{AAAI}}.

\bibitem[{Keogh and Ratanamahatana(2005)}]{keogh2005exact}
Keogh, E.; and Ratanamahatana, C.~A. 2005.
\newblock Exact indexing of dynamic time warping.
\newblock \emph{{Knowl. Inf. Syst.}}, 7: 358--386.

\bibitem[{Kingma(2014)}]{kingma2014auto}
Kingma, D.~P. 2014.
\newblock Auto-encoding variational bayes.
\newblock In \emph{{ICLR}}.

\bibitem[{Kipf and Welling(2017)}]{kipf2017semi}
Kipf, T.~N.; and Welling, M. 2017.
\newblock Semi-supervised classification with graph convolutional networks.
\newblock In \emph{{ICLR}}.

\bibitem[{Lesty{\'a}n, {\'A}cs, and Bicz{\'o}k(2022)}]{lestyan2022search}
Lesty{\'a}n, S.; {\'A}cs, G.; and Bicz{\'o}k, G. 2022.
\newblock In search of lost utility: Private location data.
\newblock In \emph{{PETS}}.

\bibitem[{Li et~al.(2016)Li, Bao, Li, Wu, Gong, and Zheng}]{li2016mining}
Li, Y.; Bao, J.; Li, Y.; Wu, Y.; Gong, Z.; and Zheng, Y. 2016.
\newblock Mining the Most Influential k-Location Set from Massive Trajectories.
\newblock In \emph{{SIGSPATIAL}}.

\bibitem[{Liu et~al.(2024)Liu, Song, Zhou, Yu, Chen, Feng, and
  Song}]{liu2024interaction}
Liu, S.; Song, J.; Zhou, Y.; Yu, N.; Chen, K.; Feng, Z.; and Song, M. 2024.
\newblock Interaction pattern disentangling for multi-agent reinforcement
  learning.
\newblock \emph{{IEEE Trans. Pattern Anal. Mach. Intell.}}, 46(12): 8157--8172.

\bibitem[{Mao et~al.(2022)Mao, Li, Li, Bai, and Zhao}]{mao2022jointly}
Mao, Z.; Li, Z.; Li, D.; Bai, L.; and Zhao, R. 2022.
\newblock Jointly contrastive representation learning on road network and
  trajectory.
\newblock In \emph{{CIKM}}.

\bibitem[{Ouyang et~al.(2018)Ouyang, Shokri, Rosenblum, and
  Yang}]{ouyang2018non}
Ouyang, K.; Shokri, R.; Rosenblum, D.~S.; and Yang, W. 2018.
\newblock A non-parametric generative model for human trajectories.
\newblock In \emph{{IJCAI}}.

\bibitem[{Radford et~al.(2018)Radford, Narasimhan, Salimans, Sutskever
  et~al.}]{radford2018improving}
Radford, A.; Narasimhan, K.; Salimans, T.; Sutskever, I.; et~al. 2018.
\newblock Improving language understanding by generative pre-training.
\newblock \emph{OpenAI blog}.

\bibitem[{Reich et~al.(2019)Reich, Budka, Robbins, and
  Hulbert}]{reich2019survey}
Reich, T.; Budka, M.; Robbins, D.; and Hulbert, D. 2019.
\newblock Survey of ETA prediction methods in public transport networks.
\newblock \emph{arXiv preprint arXiv:1904.05037}.

\bibitem[{Sanders and Schulz(2013)}]{sanders2013think}
Sanders, P.; and Schulz, C. 2013.
\newblock Think locally, act globally: Highly balanced graph partitioning.
\newblock In \emph{International Symposium on Experimental Algorithms}.

\bibitem[{Schestakov, Heinemeyer, and Demidova(2023)}]{schestakov2023road}
Schestakov, S.; Heinemeyer, P.; and Demidova, E. 2023.
\newblock Road network representation learning with vehicle trajectories.
\newblock In \emph{{PAKDD}}.

\bibitem[{Shaw, Uszkoreit, and Vaswani(2018)}]{shaw2018self}
Shaw, P.; Uszkoreit, J.; and Vaswani, A. 2018.
\newblock Self-attention with relative position representations.
\newblock In \emph{{NAACL}}.

\bibitem[{Song et~al.(2010)Song, Koren, Wang, and
  Barab{\'a}si}]{song2010modelling}
Song, C.; Koren, T.; Wang, P.; and Barab{\'a}si, A.-L. 2010.
\newblock Modelling the scaling properties of human mobility.
\newblock \emph{{Nat. Phys.}}, 6(10): 818--823.

\bibitem[{Sun et~al.(2020)Sun, Qian, Chen, Liang, Nguyen, and Yin}]{sun2020go}
Sun, K.; Qian, T.; Chen, T.; Liang, Y.; Nguyen, Q. V.~H.; and Yin, H. 2020.
\newblock Where to go next: Modeling long-and short-term user preferences for
  point-of-interest recommendation.
\newblock In \emph{{AAAI}}.

\bibitem[{Veli{\v{c}}kovi{\'c} et~al.(2018)Veli{\v{c}}kovi{\'c}, Cucurull,
  Casanova, Romero, Lio, and Bengio}]{velivckovic2018graph}
Veli{\v{c}}kovi{\'c}, P.; Cucurull, G.; Casanova, A.; Romero, A.; Lio, P.; and
  Bengio, Y. 2018.
\newblock Graph attention networks.
\newblock In \emph{{ICLR}}.

\bibitem[{Wang et~al.(2023)Wang, Wu, Zhang, Zhou, Sun, and Fu}]{wang2023human}
Wang, D.; Wu, L.; Zhang, D.; Zhou, J.; Sun, L.; and Fu, Y. 2023.
\newblock Human-instructed deep hierarchical generative learning for automated
  urban planning.
\newblock In \emph{{AAAI}}.

\bibitem[{Wang et~al.(2019)Wang, Lee, Fu, and Yu}]{wang2019learning}
Wang, M.-x.; Lee, W.-C.; Fu, T.-y.; and Yu, G. 2019.
\newblock Learning embeddings of intersections on road networks.
\newblock In \emph{{SIGSPATIAL}}.

\bibitem[{Wang et~al.(2020)Wang, Lee, Fu, and Yu}]{wang2020representation}
Wang, M.-X.; Lee, W.-C.; Fu, T.-Y.; and Yu, G. 2020.
\newblock On representation learning for road networks.
\newblock \emph{{ACM Trans. Intell. Syst. Technol.}}, 12(1): 1--27.

\bibitem[{Wang et~al.(2021)Wang, Liu, Lu, and Yang}]{wang2021large}
Wang, X.; Liu, X.; Lu, Z.; and Yang, H. 2021.
\newblock Large scale GPS trajectory generation using map based on two stage
  GAN.
\newblock \emph{{J. Data Sci.}}, 19(1): 126--141.

\bibitem[{Wang et~al.(2024{\natexlab{a}})Wang, Cao, Huang, Liu, Zheng, and
  Song}]{wang2024stega}
Wang, Y.; Cao, J.; Huang, W.; Liu, Z.; Zheng, T.; and Song, M.
  2024{\natexlab{a}}.
\newblock Spatiotemporal gated traffic trajectory simulation with
  semantic-aware graph learning.
\newblock \emph{{Inf. Fusion}}, 108: 102404.

\bibitem[{Wang et~al.(2024{\natexlab{b}})Wang, Zheng, Liang, Liu, and
  Song}]{wang2024cola}
Wang, Y.; Zheng, T.; Liang, Y.; Liu, S.; and Song, M. 2024{\natexlab{b}}.
\newblock Cola: Cross-city mobility transformer for human trajectory
  simulation.
\newblock In \emph{{WWW}}.

\bibitem[{Wang et~al.(2024{\natexlab{c}})Wang, Zheng, Liu, Feng, Chen, Hao, and
  Song}]{wang2024star}
Wang, Y.; Zheng, T.; Liu, S.; Feng, Z.; Chen, K.; Hao, Y.; and Song, M.
  2024{\natexlab{c}}.
\newblock Spatiotemporal-augmented graph neural networks for human mobility
  simulation.
\newblock \emph{{IEEE Trans. Knowl. Data Eng.}}, 36(11): 7074–7086.

\bibitem[{Wen et~al.(2024)Wen, Lin, Wu, Mao, Cai, Hou, Guo, Liang, Jin, Zhao,
  Zimmermann, Ye, and Wan}]{wen2024survey}
Wen, H.; Lin, Y.; Wu, L.; Mao, X.; Cai, T.; Hou, Y.; Guo, S.; Liang, Y.; Jin,
  G.; Zhao, Y.; Zimmermann, R.; Ye, J.; and Wan, H. 2024.
\newblock A survey on service route and time prediction in instant delivery:
  Taxonomy, progress, and prospects.
\newblock \emph{{IEEE Trans. Knowl. Data Eng.}}, 36(12): 7516--7535.

\bibitem[{Wu et~al.(2020)Wu, Zhao, Wang, and Pan}]{wu2020learning}
Wu, N.; Zhao, X.~W.; Wang, J.; and Pan, D. 2020.
\newblock Learning effective road network representation with hierarchical
  graph neural networks.
\newblock In \emph{{SIGKDD}}.

\bibitem[{Xie, Li, and Phillips(2017)}]{xie2017distributed}
Xie, D.; Li, F.; and Phillips, J.~M. 2017.
\newblock Distributed trajectory similarity search.
\newblock In \emph{{VLDB}}.

\bibitem[{Xu et~al.(2020)Xu, Ruan, Korpeoglu, Kumar, and
  Achan}]{xu2020inductive}
Xu, D.; Ruan, C.; Korpeoglu, E.; Kumar, S.; and Achan, K. 2020.
\newblock Inductive representation learning on temporal graphs.
\newblock In \emph{{ICLR}}.

\bibitem[{Yang and Gidofalvi(2018)}]{yang2018fast}
Yang, C.; and Gidofalvi, G. 2018.
\newblock Fast map matching, an algorithm integrating hidden Markov model with
  precomputation.
\newblock \emph{{Int. J. Geogr. Inf. Sci.}}, 32(3): 547--570.

\bibitem[{Yang et~al.(2020)Yang, Fankhauser, Rosso, and
  Cudre-Mauroux}]{yang2020location}
Yang, D.; Fankhauser, B.; Rosso, P.; and Cudre-Mauroux, P. 2020.
\newblock Location prediction over sparse user mobility traces using RNNs:
  Flashback in hidden states!
\newblock In \emph{{IJCAI}}.

\bibitem[{Yu et~al.(2017)Yu, Zhang, Wang, and Yu}]{yu2017seqgan}
Yu, L.; Zhang, W.; Wang, J.; and Yu, Y. 2017.
\newblock Seqgan: Sequence generative adversarial nets with policy gradient.
\newblock In \emph{{AAAI}}.

\bibitem[{Yuan, Li, and Bao(2022)}]{yuan2022route}
Yuan, H.; Li, G.; and Bao, Z. 2022.
\newblock Route travel time estimation on a road network revisited:
  Heterogeneity, proximity, periodicity and dynamicity.
\newblock In \emph{{VLDB}}.

\bibitem[{Yuan et~al.(2010)Yuan, Zheng, Zhang, Xie, Xie, Sun, and
  Huang}]{yuan2010tdrive}
Yuan, J.; Zheng, Y.; Zhang, C.; Xie, W.; Xie, X.; Sun, G.; and Huang, Y. 2010.
\newblock T-drive: Driving directions based on taxi trajectories.
\newblock In \emph{{SIGSPATIAL}}.

\bibitem[{Zhang and Long(2023)}]{zhang2023road}
Zhang, L.; and Long, C. 2023.
\newblock Road network representation learning: A dual graph-based approach.
\newblock \emph{{ACM Trans. Knowl. Discov. Data}}, 17(9): 1--25.

\bibitem[{Zhao et~al.(2022)Zhao, Huang, Wang, Chai, and Li}]{zhao2022rne}
Zhao, T.; Huang, S.; Wang, Y.; Chai, C.; and Li, G. 2022.
\newblock RNE: Computing shortest paths using road network embedding.
\newblock \emph{{VLDB J.}}, 31(3): 507--528.

\bibitem[{Zheng et~al.(2022)Zheng, Feng, Zhang, Hao, Song, Wang, Wang, Zhao,
  and Chen}]{zheng2022transition}
Zheng, T.; Feng, Z.; Zhang, T.; Hao, Y.; Song, M.; Wang, X.; Wang, X.; Zhao,
  J.; and Chen, C. 2022.
\newblock Transition propagation graph neural networks for temporal networks.
\newblock \emph{{IEEE Trans. Neural Networks Learn. Syst.}}, 35(4): 4567--4579.

\bibitem[{Zheng et~al.(2023)Zheng, Wang, Feng, Song, Hao, Song, Wang, Wang, and
  Chen}]{zheng2023temporal}
Zheng, T.; Wang, X.; Feng, Z.; Song, J.; Hao, Y.; Song, M.; Wang, X.; Wang, X.;
  and Chen, C. 2023.
\newblock Temporal aggregation and propagation graph neural networks for
  dynamic representation.
\newblock \emph{{IEEE Trans. Knowl. Data Eng.}}, 35(10): 10151--10165.

\bibitem[{Zheng(2015)}]{zheng2015trajectory}
Zheng, Y. 2015.
\newblock Trajectory data mining: An overview.
\newblock \emph{{ACM Trans. Intell. Syst. Technol.}}, 6(3): 1--41.

\bibitem[{Zhu et~al.(2023{\natexlab{a}})Zhu, Ye, Wu, Zhao, and
  Yu}]{zhu2023synmob}
Zhu, Y.; Ye, Y.; Wu, Y.; Zhao, X.; and Yu, J. 2023{\natexlab{a}}.
\newblock SynMob: creating high-fidelity synthetic GPS trajectory dataset for
  urban mobility analysis.
\newblock In \emph{{NeurIPS}}.

\bibitem[{Zhu et~al.(2023{\natexlab{b}})Zhu, Ye, Zhang, Zhao, and
  Yu}]{zhu2023difftraj}
Zhu, Y.; Ye, Y.; Zhang, S.; Zhao, X.; and Yu, J. 2023{\natexlab{b}}.
\newblock DiffTraj: generating GPS trajectory with diffusion probabilistic
  model.
\newblock In \emph{{NeurIPS}}.

\bibitem[{Zhu et~al.(2024)Zhu, Yu, Zhao, Liu, Ye, Chen, Zhang, Wei, and
  Liang}]{zhu2024controltraj}
Zhu, Y.; Yu, J.~J.; Zhao, X.; Liu, Q.; Ye, Y.; Chen, W.; Zhang, Z.; Wei, X.;
  and Liang, Y. 2024.
\newblock Controltraj: Controllable trajectory generation with
  topology-constrained diffusion model.
\newblock In \emph{{SIGKDD}}.

\end{thebibliography}

\clearpage
\appendix

\section{Details of HOSER Framework}

\subsection{Details of Metric Features}

In the Destination-Oriented Navigator module, we incorporate metric information from the candidate road segment $r_\text{c}$ to the destination $r_\text{dest}$ to model destination guidance. Specifically, we consider the following two primary metrics:

\begin{enumerate}
\item The distance from the candidate road segment $r_\text{c}$ to the destination $r_\text{dest}$, denoted as $d(r_\text{c}, r_\text{dest})$.
\item The angle between the direction of $r_\text{c}$ and the line connecting the current position to the destination $r_\text{dest}$, denoted as $\phi(r_\text{c}, r_\text{dest})$.
\end{enumerate}

This approach is based on the observation that humans, when navigating, tend to choose routes that are close to their destination and aligned with its direction. To effectively utilize neural networks for learning from these two entities, we first normalize them, yielding $\hat{d}(r_\text{c}, r_\text{dest})$ and $\hat{\phi}(r_\text{c}, r_\text{dest})$:
\begin{equation}
\resizebox{.875\columnwidth}{!}{$
\left\{
\begin{aligned}
\hat{d}(r_\text{c}, r_\text{dest}) &= \logonep \big(d(r_\text{c}, r_\text{dest}) - \min_{r_\text{c}^\prime \in R(r_i)}\big\{d(r_\text{c}^\prime, r_\text{dest})\big\}\big) \\
\hat{\phi}(r_\text{c}, r_\text{dest}) &= \frac{1}{\pi} \phi(r_\text{c}, r_\text{dest})
\end{aligned}
\right.
$}
\end{equation}
Subsequently, we apply learnable linear transformations to the aforementioned two metrics, resulting in $\boldsymbol{h}_{r_\text{c}, r_\text{dest}} \in \mathbb{R}^{2d}$, which can be written as:
\begin{equation}
\boldsymbol{h}_{r_\text{c}, r_\text{dest}} = \hat{d}(r_\text{c}, r_\text{dest}) \theta_d \ \Vert \ \hat{\phi}(r_\text{c}, r_\text{dest}) \theta_\phi,
\end{equation}
where $\theta_d, \theta_\phi \in \mathbb{R}^d$ are learnable parameters used for projection and ``$\Vert$'' is the concatenation operation.

\subsection{Optimal Trajectory Searching Algorithm}

For the conditional information $(r_\text{org}, t_\text{org}, r_\text{dest})$ provided, we choose the trajectory with the highest probability as the final generated trajectory. In practice, we first convert the probabilistic representation of human movement policy into its negative logarithmic form. This transformation enables the original problem of maximizing the cumulative trajectory probability, expressed as a product, to be reformulated as minimizing the cumulative sum of the corresponding negative logarithms, as shown below:
\begin{equation}
\resizebox{.875\columnwidth}{!}{$
\begin{aligned}
\max \prod_{i=1}^{n-1}\pi(a_i, s_i) = &\max \prod_{i=1}^{n-1}P(r_{i+1} \mid x_{1:i}, r_\text{dest}) \\
\iff &\min \sum_{i=1}^{n-1} -\log P(r_{i+1} \mid x_{1:i}, r_\text{dest}), \\
\mathrm{s.t.} \quad x_1 &= (r_\text{org}, t_\text{org}), \quad r_n = r_\text{dest}.
\end{aligned}
$}
\end{equation}
Then we utilize a min-heap to efficiently retrieve the next candidate road segment for processing based on the highest known probability, as demonstrated in \Cref{alg:optimal_trajectory_search}.

\begin{algorithm}[ht]
\footnotesize
\SetAlgoLined
\SetKwInOut{Input}{Input}
\SetKwInOut{Output}{Output}
\Input{Road network $\mathcal{G} = \langle \mathcal{V}, \mathcal{E} \rangle$,\\
conditional information $(r_{\mathrm{org}}, t_{\mathrm{org}}, r_{\mathrm{dest}})$.}
\Output{A synthetic trajectory $[x_1, x_2, \ldots, x_n]$ such that $x_1 = (r_{\mathrm{org}}, t_{\mathrm{org}})$ and $r_n = r_{\mathrm{dest}}$.}

Initialize a min-heap $H \gets \varnothing$\;
Initialize $\Phi[r] \gets +\infty$ for all road segments $r$\;
Initialize $S[r] \gets \varnothing$ for all road segments $r$\;
$\Phi[r_{\mathrm{org}}] \gets 0$, $S[r_{\mathrm{org}}] \gets [(r_{\mathrm{org}}, t_{\mathrm{org}})]$\;
Insert $(\Phi[r_{\mathrm{org}}], r_{\mathrm{org}})$ into $H$\;

\While{$H$ is not empty}{
    $(negLogProbSum, r) \gets \textsc{GetHeapTop}(H)$\;
    \lIf{$r = r_{\mathrm{dest}}$}{\KwRet{$S[r]$}}
    \lIf{$negLogProbSum > \Phi[r]$}{\textbf{continue}}
    \ForEach{$r^\prime \in R(r)$}{
        Predict $P_{\theta}(r^\prime \mid S[r], r_{\mathrm{dest}})$ and timestamp $t$\;
        \If{$\Phi[r^\prime] > \Phi[r] - \log P_{\theta}(r^\prime \mid S[r], r_{\mathrm{dest}})$}{
            $\Phi[r^\prime] \gets \Phi[r] - \log P_{\theta}(r^\prime \mid S[r], r_{\mathrm{dest}})$\;
            $S[r^\prime] \gets S[r] + [(r^\prime, t)]$\;
            Insert $(\Phi[r^\prime], r^\prime)$ into $H$\;
        }
    }
}

\caption{Optimal Trajectory Search}
\label{alg:optimal_trajectory_search}
\end{algorithm}

\section{Details of the Experimental Setup}

In this section, we provide a detailed description of the experimental setup used in this paper, including the dataset, evaluation metrics, and baseline methods.

\subsection{Datasets}

We evaluate the performance of HOSER and other baselines on trajectory datasets from three different cities: Beijing, Porto\footnote{\url{https://www.kaggle.com/c/pkdd-15-taxi-trip-time-prediction-ii/data}}, and San Francisco\footnote{\url{https://ieee-dataport.org/open-access/crawdad-epflmobility}}. Due to the original dataset containing GPS trajectories, we collect the road networks of the three cities above from OpenStreetMap\footnote{\url{https://www.openstreetmap.org}} and perform map matching~\cite{yang2018fast} to convert the GPS sequences in the original dataset into road sequences. For all datasets, we filter trajectories with lengths shorter than 5, containing loops, or exhibiting time intervals greater than 15 minutes. The statistics of the three datasets after preprocessing are shown in \Cref{tab:appendix_statistics_of_datasets}.

\begin{table}[ht]
\centering
\small
\begin{tabular}{lccc}
\toprule
Data statistics & Beijing & Porto & San Francisco \\
\midrule
Roads Number & \num{40060} & \num{11024} & \num{27187} \\
Trajectory Number & \num{899115} & \num{687656} & \num{293023} \\
Average Distance & 5.16km & 3.66km & 3.41km \\
Average Time & 12.87min & 7.99min & 9.29min \\
\bottomrule
\end{tabular}
\caption{Statistics of three real-world trajectory datasets.}
\label{tab:appendix_statistics_of_datasets}
\end{table}

\subsection{Evaluation Metrics}

To accurately evaluate the similarity between the generated trajectories and the real trajectories, we adopt the widely-used evaluation setup from previous studies~\cite{jiang2023continuous, wang2024stega}, assessing the generated trajectories from both global and local perspectives.

From a global perspective, we focus on the overall distribution of the trajectories using the following three metrics: 1) \textit{Distance}, which measures the travel distance of the trajectory; 2) \textit{Radius}, which calculates the radius of gyration of the trajectory~\cite{gonzalez2008understanding}; and 3) \textit{Duration}~\cite{feng2020learning}, which counts the dwell duration among locations. To obtain quantitative results, we use Jensen-Shannon divergence (JSD) to assess the similarity between the distributions of these three metrics. Let $P$ denote the probability distribution of the real trajectories and $Q$ denote the probability distribution of the generated trajectories. The JSD is then calculated as follows:
\begin{equation}
\resizebox{.875\columnwidth}{!}{$
\mathrm{JSD}(P \ \Vert \ Q) = \frac{1}{2}\mathbb{E}_P \big[\log \frac{2P}{P + Q}\big] + \frac{1}{2}\mathbb{E}_Q \big[\log \frac{2Q}{P + Q}\big].
$}
\end{equation}
In the implementation, the aforementioned metrics are analyzed by dividing their values into histogram bins.  The upper bound of each metric is determined by the maximum value observed in the real data, while the lower bound is set to $0$.  The range between these bounds is uniformly divided into $100$ intervals, enabling a comparison of the distributions of real and generated trajectories for a given metric.

From a local perspective, we first divide the city into a series of $200\text{m} \times 200\text{m}$ grids, and then compare the similarity between real trajectories and generated trajectories that share the same origin-destination (OD) grids, using the following three metrics: 1) \textit{Hausdorff distance}\cite{xie2017distributed}, which measures the maximum distance between the spatio-temporal points of two trajectories; 2) \textit{DTW}\cite{keogh2005exact}, which calculates the similarity between two trajectories by optimally aligning them; and 3) \textit{EDR}~\cite{chen2005robust}, which measures the minimum number of edits required to transform one trajectory into the other.

\subsection{Baselines}

\begin{itemize}
\item \textbf{Markov}~\cite{gambs2012next}: The Markov method is a simple yet efficient probabilistic approach that employs Markov chains to describe transitions between states. It treats road segments as states and uses trajectories to estimate the transition probabilities between these road segments.
\item \textbf{Dijkstra's algorithm}~\cite{dijkstra1959note}: In this method, we achieve trajectory generation by finding the shortest path between a given OD pair.
\item \textbf{SeqGAN}~\cite{yu2017seqgan}: This model leverages GAN for sequence generation by framing the generator as a stochastic policy within a reinforcement learning framework and utilizing the discriminator's output as a reward.
\item \textbf{SVAE}~\cite{huang2019variational}: In this method, the VAE and Sequence to Sequence (Seq2Seq) models are combined to achieve trajectory generation.
\item \textbf{MoveSim}~\cite{feng2020learning}: MoveSim is a GAN-based trajectory generation method that incorporates the physical principles of human movement.
\item \textbf{TSG}~\cite{wang2021large}: TSG is a two-stage GAN-based approach. In the first stage, the method divides the map into multiple grids and generates coarse-grained trajectories at the grid level. In the second stage, these initial trajectories are further refined utilizing map images to achieve finer detail.
\item \textbf{TrajGen}~\cite{cao2021generating}: This method maps the trajectory data to a spatial grid and employs a Seq2Seq model to generate the trajectory data.
\item \textbf{DiffTraj}~\cite{zhu2023difftraj}: DiffTraj is a diffusion-based trajectory generation method, which first adds noise to the trajectory data and then progressively removes the noise to generate trajectories.
\item \textbf{STEGA}~\cite{wang2024stega}: STEGA devises two spatio-temporal gates equipped with semantic-aware graph learning for continuous trajectory generation.
\item \textbf{TS-TrajGen}~\cite{jiang2023continuous}: TS-TrajGen achieves trajectory generation by combining a two-stage GAN method with the A* algorithm.
\end{itemize}

\section{Additional Experiments}

\subsection{Overall Performance}

\paragraph{Quantitative Analysis.}

Due to space constraints, the DTW and EDR metrics (both at the local level) for the three real-world trajectory datasets are presented in \Cref{tab:appendix_performance_comparison}. We can observe that HOSER surpasses other \emph{state-of-the-art} baselines on these two metrics, further validating its exceptional ability to generate high-fidelity trajectories.

\begin{table*}[t]
\centering
\small
\begin{tabular}{lcccccc}
\toprule
& \multicolumn{2}{c|}{Beijing} & \multicolumn{2}{c|}{Porto} & \multicolumn{2}{c}{San Francisco} \\
\cmidrule(lr){2-7}
Methods & DTW($\mathrel{\downarrow}$) & \multicolumn{1}{c|}{EDR($\mathrel{\downarrow}$)} & DTW($\mathrel{\downarrow}$) & \multicolumn{1}{c|}{EDR($\mathrel{\downarrow}$)} & DTW($\mathrel{\downarrow}$) & EDR($\mathrel{\downarrow}$) \\
\midrule
Markov & 13.3794 & \multicolumn{1}{c|}{0.6244} & 18.4359 & \multicolumn{1}{c|}{0.7019} & 18.7161 & 0.7718 \\
Dijkstra & 9.7251 & \multicolumn{1}{c|}{0.5379} & \underline{14.8005} & \multicolumn{1}{c|}{0.6051} & \underline{11.1584} & \underline{0.7044} \\
SeqGAN & 10.9190 & \multicolumn{1}{c|}{0.5422} & 18.8179 & \multicolumn{1}{c|}{0.6237} & 17.2669 & 0.6937 \\
SVAE & 9.8810 & \multicolumn{1}{c|}{0.8622} & 15.2379 & \multicolumn{1}{c|}{0.8861} & 12.0000 & 0.8240 \\
MoveSim & 61.8770 & \multicolumn{1}{c|}{0.8444} & 34.8874 & \multicolumn{1}{c|}{0.8929} & 34.1062 & 0.9120 \\
TSG & 21.1353 & \multicolumn{1}{c|}{0.8913} & 19.5944 & \multicolumn{1}{c|}{0.9051} & 39.6601 & 0.9010 \\
TrajGen & 60.2799 & \multicolumn{1}{c|}{0.8382} & 26.2214 & \multicolumn{1}{c|}{0.8747} & 51.7043 & 0.8774 \\
DiffTraj & \underline{8.7593} & \multicolumn{1}{c|}{0.8125} & 14.9034 & \multicolumn{1}{c|}{0.7663} & 12.1653 & 0.8682 \\
STEGA & 11.9817 & \multicolumn{1}{c|}{0.6630} & 17.2801 & \multicolumn{1}{c|}{0.7282} & 17.4998 & 0.8035 \\
TS-TrajGen & 19.2226 & \multicolumn{1}{c|}{0.6399} & 19.2635 & \multicolumn{1}{c|}{0.6894} & 19.1613 & 0.7923 \\
\cmidrule(lr){1-7}
Markov* & 8.8718 & \multicolumn{1}{c|}{\underline{0.4640}} & 15.9415 & \multicolumn{1}{c|}{\underline{0.5488}} & 14.0063 & 0.6647 \\
SeqGAN* & 9.2388 & \multicolumn{1}{c|}{0.4669} & 18.1768 & \multicolumn{1}{c|}{0.5529} & 18.2834 & 0.6685 \\
MoveSim* & 34.6229 & \multicolumn{1}{c|}{0.6961} & 26.9302 & \multicolumn{1}{c|}{0.7218} & 20.1016 & 0.8063 \\
STEGA* & 11.4817 & \multicolumn{1}{c|}{0.4940} & 16.0648 & \multicolumn{1}{c|}{0.6673} & 15.4039 & 0.7179 \\
\cmidrule(lr){1-7}
Ours w/o RNE & 7.7484 & \multicolumn{1}{c|}{0.4604} & 13.8254 & \multicolumn{1}{c|}{0.5641} & 11.1391 & 0.6457 \\
Ours w/o TrajE & 7.5486 & \multicolumn{1}{c|}{0.4563} & 13.9354 & \multicolumn{1}{c|}{0.5506} & 11.2309 & 0.6401 \\
Ours w/o Nav & 7.8913 & \multicolumn{1}{c|}{0.4637} & 14.0036 & \multicolumn{1}{c|}{0.5673} & 12.0037 & 0.6545 \\
Ours & \textbf{7.2964} & \multicolumn{1}{c|}{\textbf{0.4483}} & \textbf{13.4834} & \multicolumn{1}{c|}{\textbf{0.5436}} & \textbf{11.0509} & \textbf{0.6319} \\
\bottomrule
\end{tabular}
\caption{Comparison of DTW and EDR metrics across three trajectory datasets. Method names marked with an asterisk (*) indicate their corresponding search versions. \textbf{Bold} and \underline{underline} indicate the best and the second best results, respectively. The symbol $\mathrel{\downarrow}$ indicates that a lower value is preferable. All results are averaged over 5 distinct random seeds (ranging from 0 to 4).}
\label{tab:appendix_performance_comparison}
\end{table*}

\paragraph{Discussion on Baselines with the Search Algorithm.}

As shown in \Cref{tab:appendix_performance_comparison}, although these modified models generally show improvements on two metrics compared to their original counterparts, their performance could not match that of HOSER. This once again verifies that it is hard to rely solely on a search-based paradigm, without holistically modeling human mobility patterns, to yield optimal results.

\paragraph{Ablation Studies.}

To investigate the impact of the three key modules in HOSER(i.e., the Road Network Encoder, the Trajectory Encoder, and the Navigator) on the effectiveness of trajectory generation, we conduct an ablation study by comparing the performance of three variants of HOSER:

\begin{itemize}
\item Ours w/o RNE: In this variant, we remove the Road Network Encoder from HOSER and rely solely on each road segment's ID for encoding.
\item Ours w/o TrajE: This variant removes the Trajectory Encoder, and relies solely on destination guidance to generate trajectories.
\item Ours w/o Nav: In this variant, the model predicts the next spatio-temporal point using only the partial trajectory, without considering destination guidance.
\end{itemize}

As shown in Table 1, our model outperforms variants with specific modules removed in both the DTW and EDR metrics, further validating that the three aforementioned modules are essential for high-quality trajectory generation. Specifically, the removal of the Navigator module has the most significant impact on the DTW and EDR metrics, highlighting the crucial role of destination guidance in trajectory generation. Additionally, eliminating either the Road Network Encoder or the Trajectory Encoder results in a decline in the DTW and EDR indicators, demonstrating the necessity of road network features and partial trajectory semantic information for effective trajectory generation.

\paragraph{Visualization Analysis.}

\begin{figure*}[t]
\centering
\subcaptionbox{Dijkstra}{\includegraphics[width=0.24\textwidth]{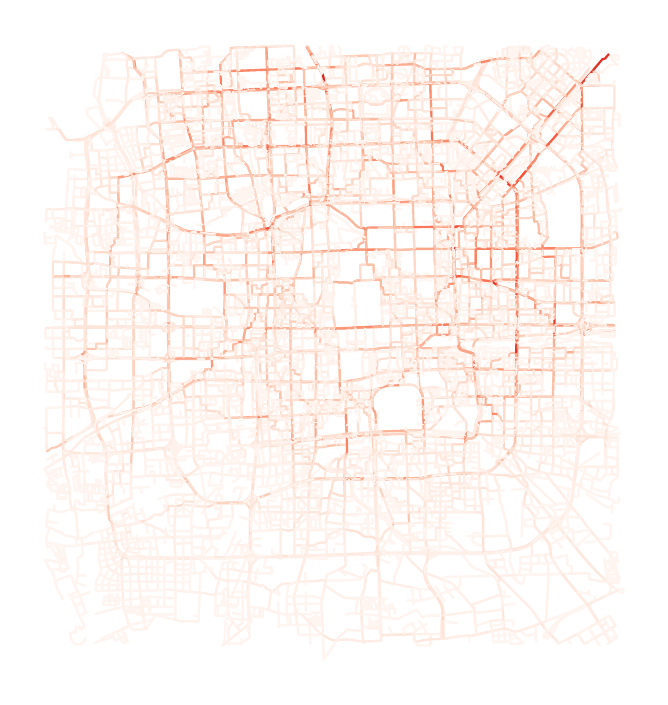}}
\subcaptionbox{DiffTraj}{\includegraphics[width=0.24\textwidth]{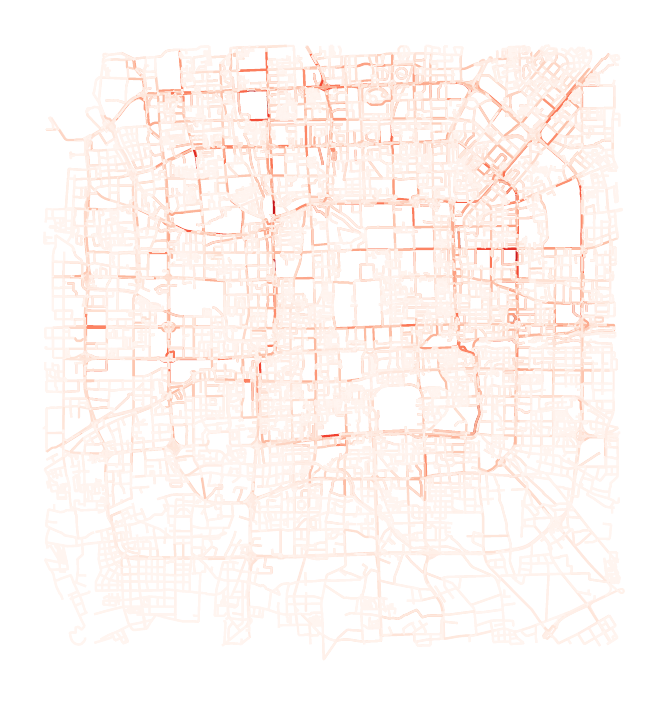}}
\subcaptionbox{Ours}{\includegraphics[width=0.24\textwidth]{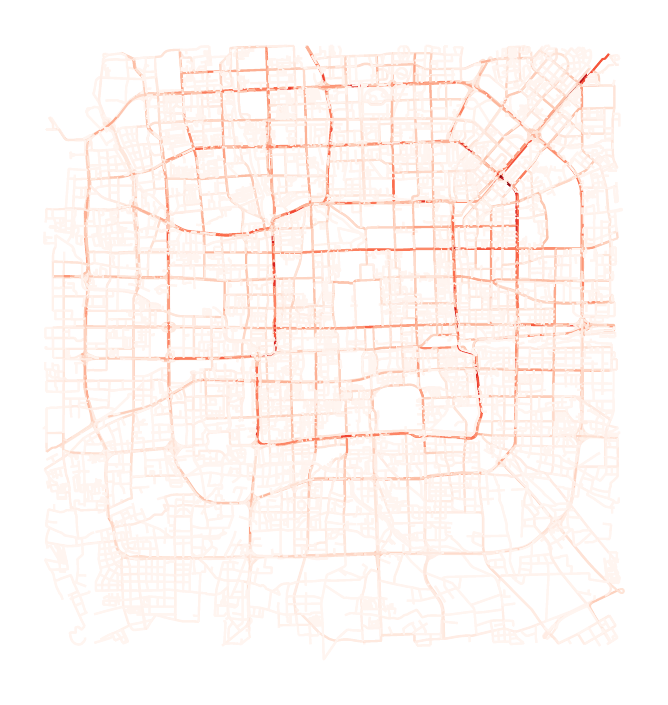}}
\subcaptionbox{Real}{\includegraphics[width=0.24\textwidth]{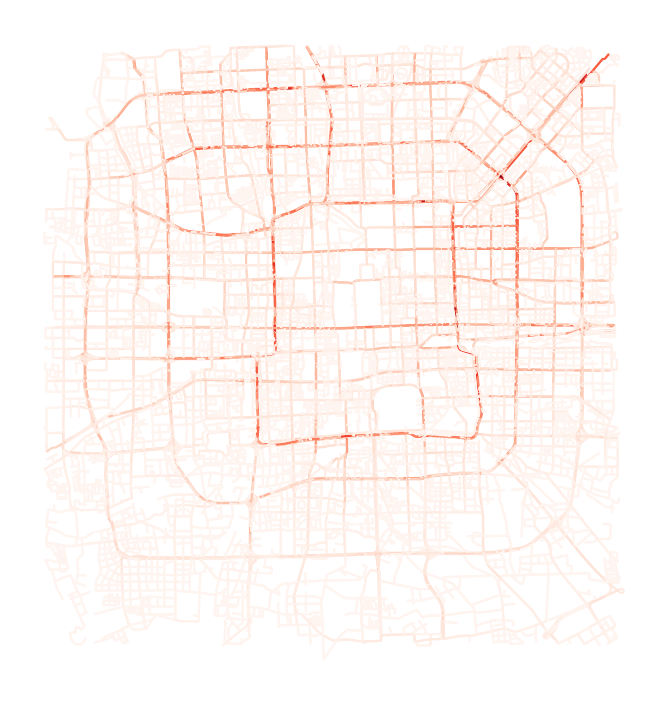}}
\caption{Visualization of the trajectories in Beijing.}
\label{fig:appendix_vis_heatmap_beijing}
\end{figure*}

\begin{figure*}[t]
\centering
\subcaptionbox{Dijkstra}{\includegraphics[width=0.24\textwidth]{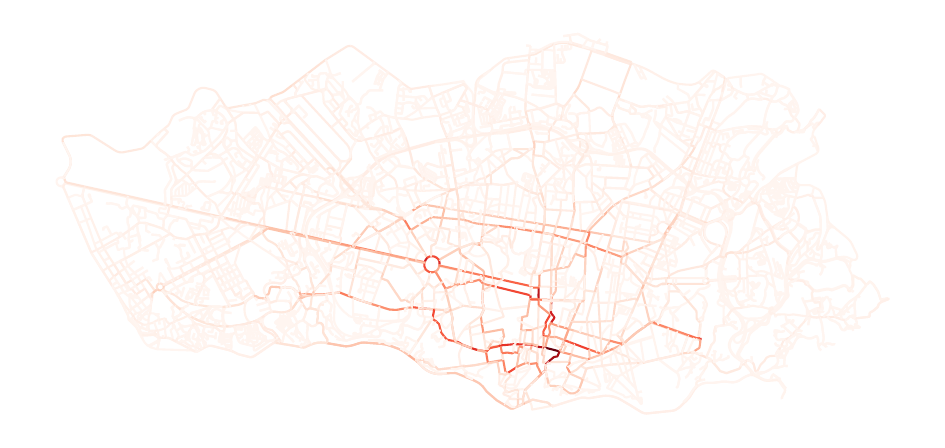}}
\subcaptionbox{DiffTraj}{\includegraphics[width=0.24\textwidth]{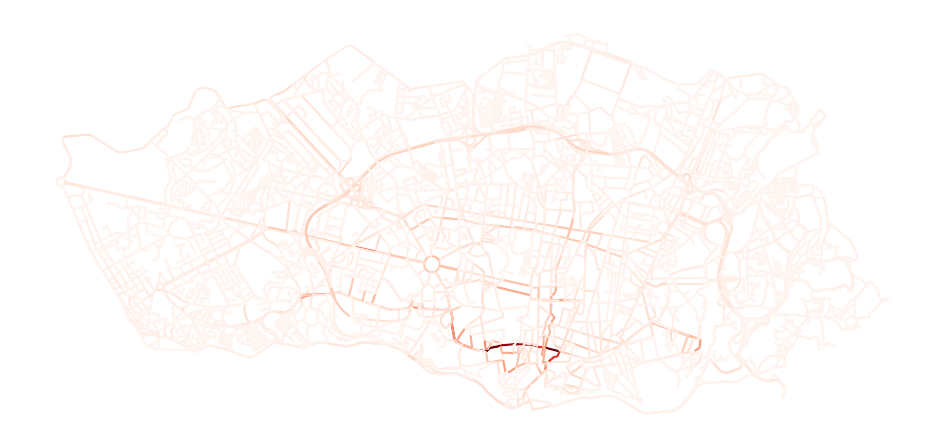}}
\subcaptionbox{Ours}{\includegraphics[width=0.24\textwidth]{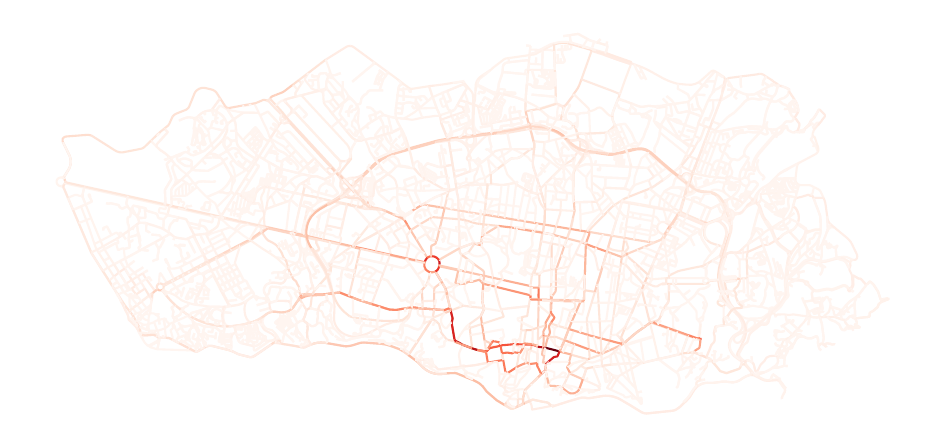}}
\subcaptionbox{Real}{\includegraphics[width=0.24\textwidth]{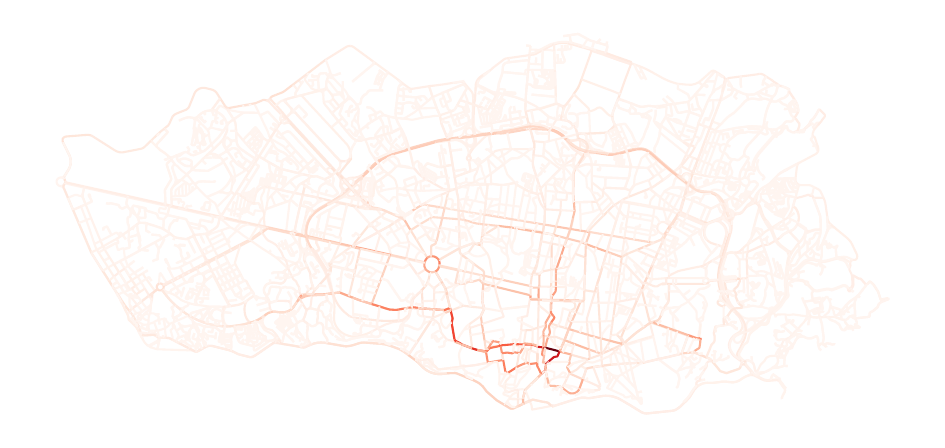}}
\caption{Visualization of the trajectories in Porto.}
\label{fig:appendix_vis_heatmap_porto}
\end{figure*}

\begin{figure*}[t]
\centering
\subcaptionbox{Dijkstra}{\includegraphics[width=0.24\textwidth]{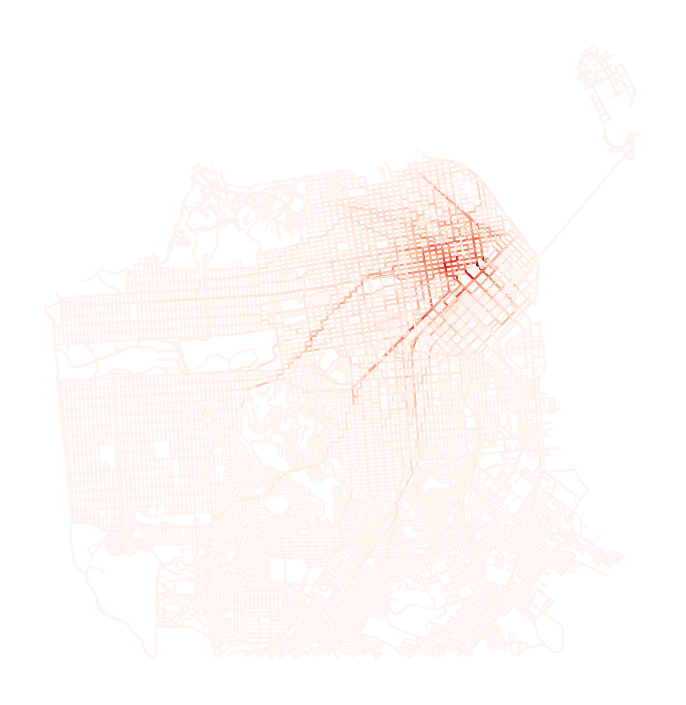}}
\subcaptionbox{DiffTraj}{\includegraphics[width=0.24\textwidth]{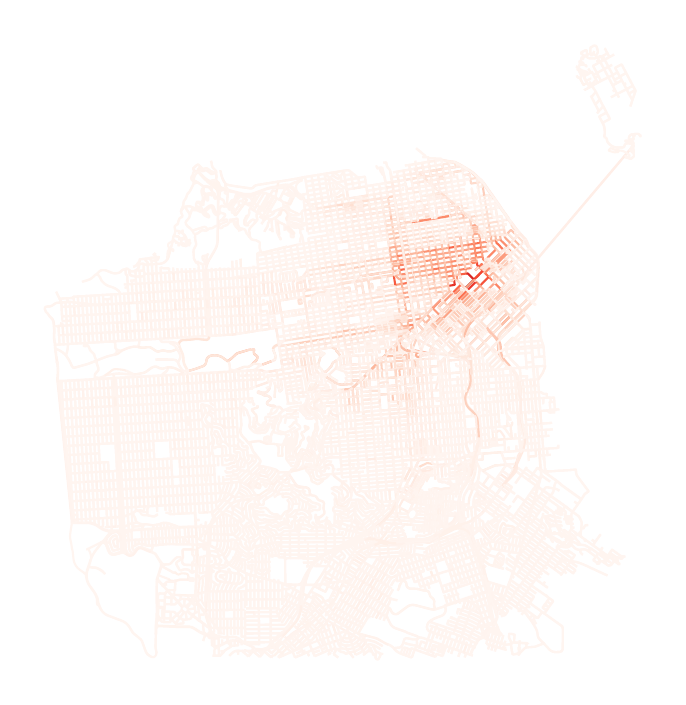}}
\subcaptionbox{Ours}{\includegraphics[width=0.24\textwidth]{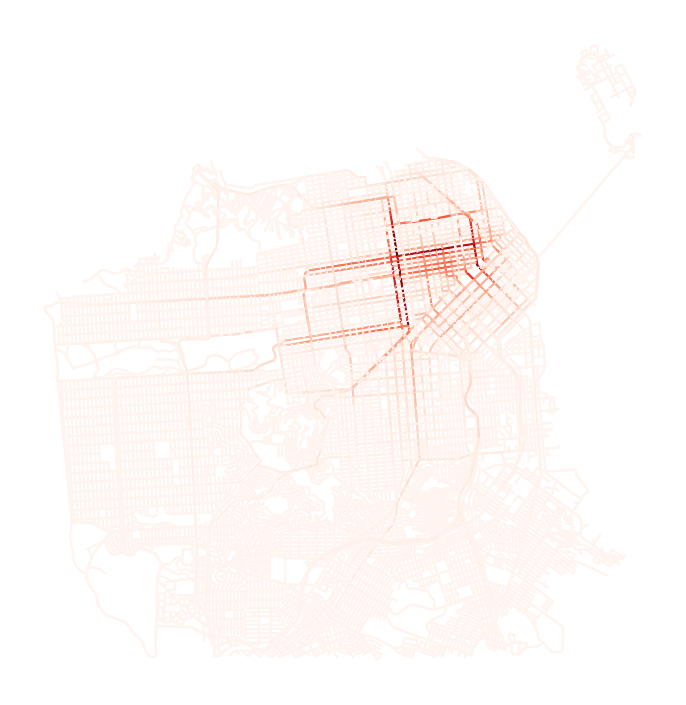}}
\subcaptionbox{Real}{\includegraphics[width=0.24\textwidth]{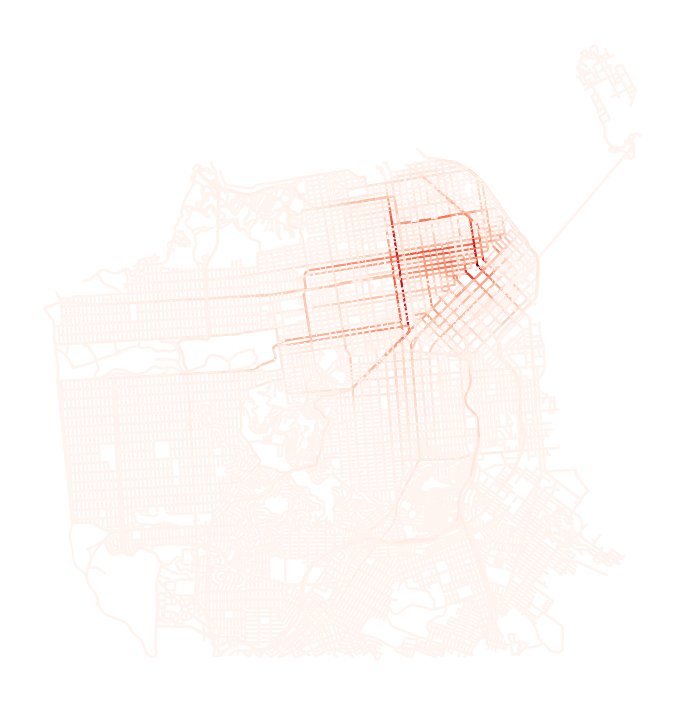}}
\caption{Visualization of the trajectories in San Francisco.}
\label{fig:appendix_vis_heatmap_san_francisco}
\end{figure*}

Due to space limitations, we present the visualization results of the trajectories generated by different methods and the corresponding ground truth trajectories in three cities, as shown in \Cref{fig:appendix_vis_heatmap_beijing}, \Cref{fig:appendix_vis_heatmap_porto}, and \Cref{fig:appendix_vis_heatmap_san_francisco} respectively. Since Dijkstra's algorithm fails to account for the complexities of human mobility patterns, and DiffTraj does not effectively utilize road network information to guide trajectory generation, the trajectories generated by these methods still exhibit certain discrepancies compared to the distribution of real trajectories. In contrast, our method holistically models human mobility patterns, accurately reproducing real trajectory distributions and significantly outperforming other methods.

\begin{figure*}[t]
\centering
\subcaptionbox{Ours}{\includegraphics[width=0.4\textwidth]{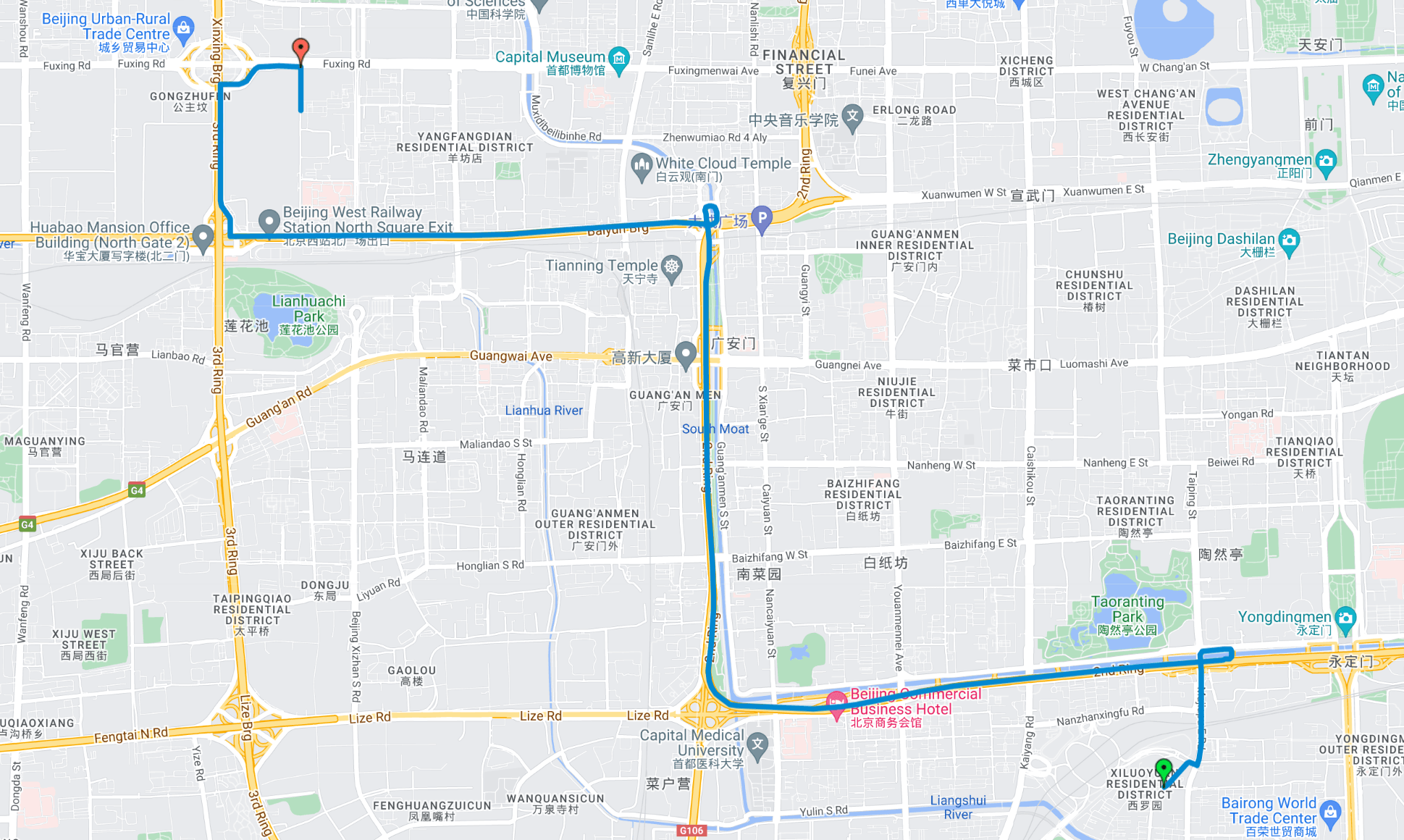}}
\subcaptionbox{Real}{\includegraphics[width=0.4\textwidth]{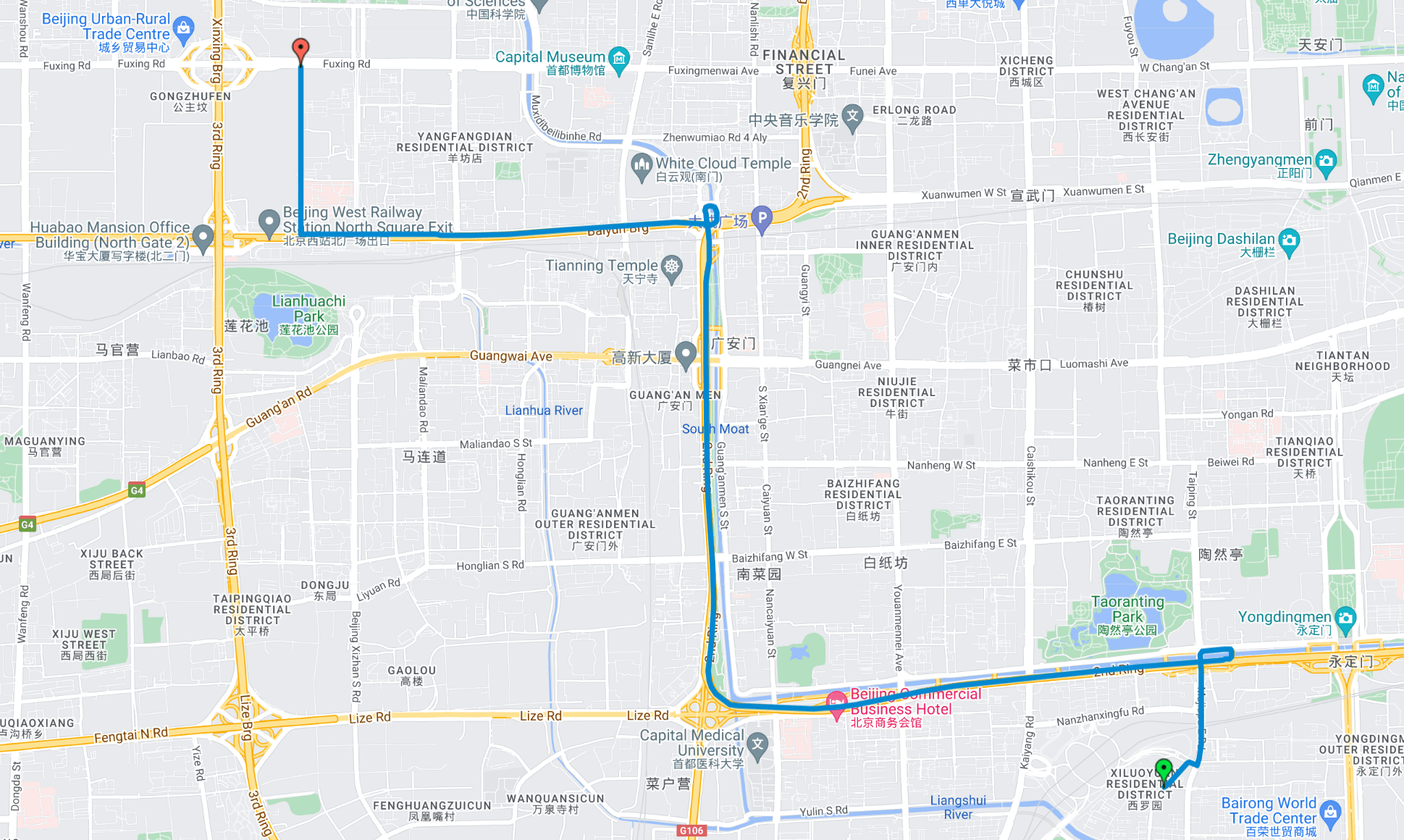}}
\caption{Visualization of the trajectory generated by HOSER and the real trajectory for the same OD pair in Beijing.}
\label{fig:appendix_vis_traj_beijing}
\end{figure*}

\begin{figure*}[t]
\centering
\subcaptionbox{Ours}{\includegraphics[width=0.4\textwidth]{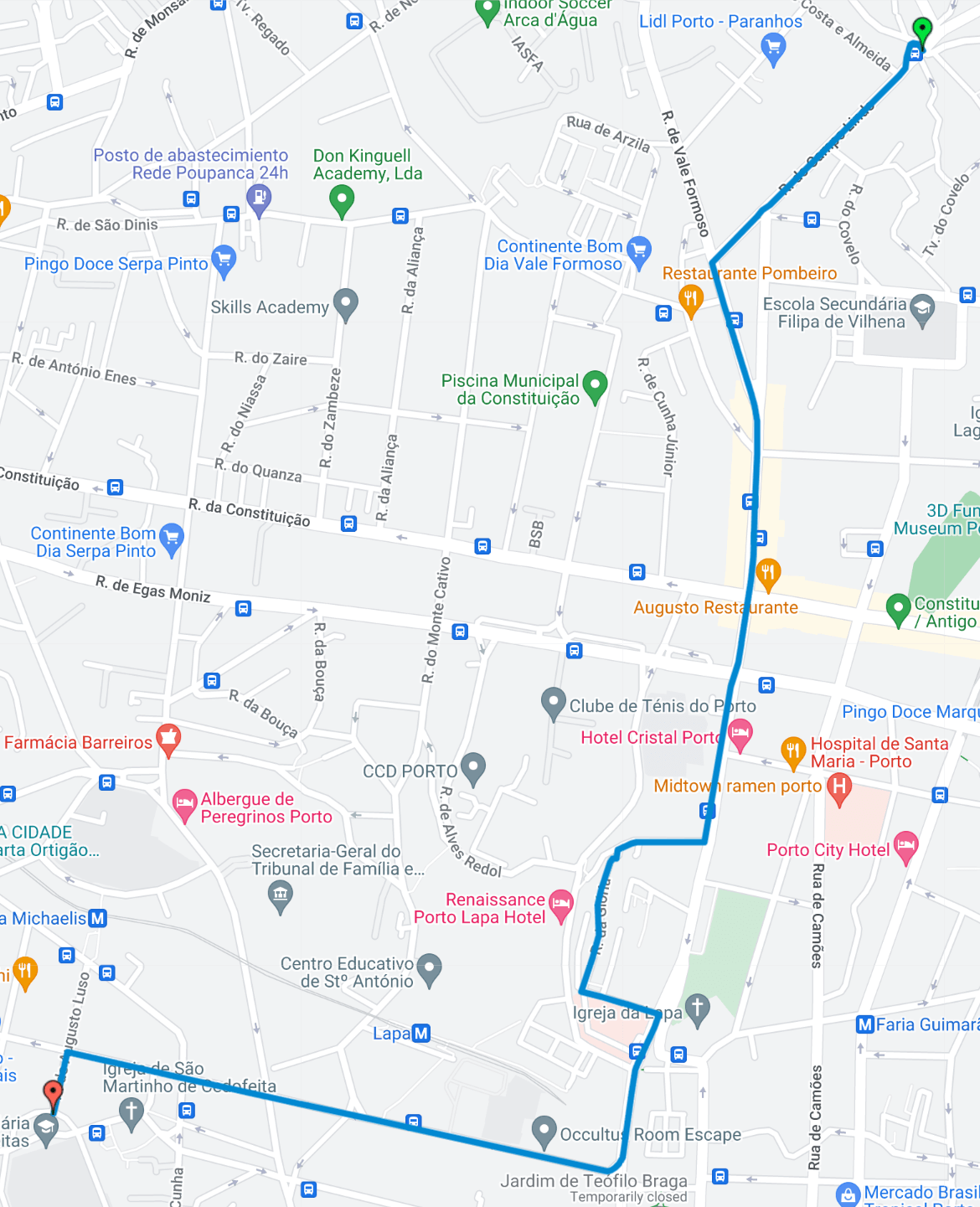}}
\subcaptionbox{Real}{\includegraphics[width=0.4\textwidth]{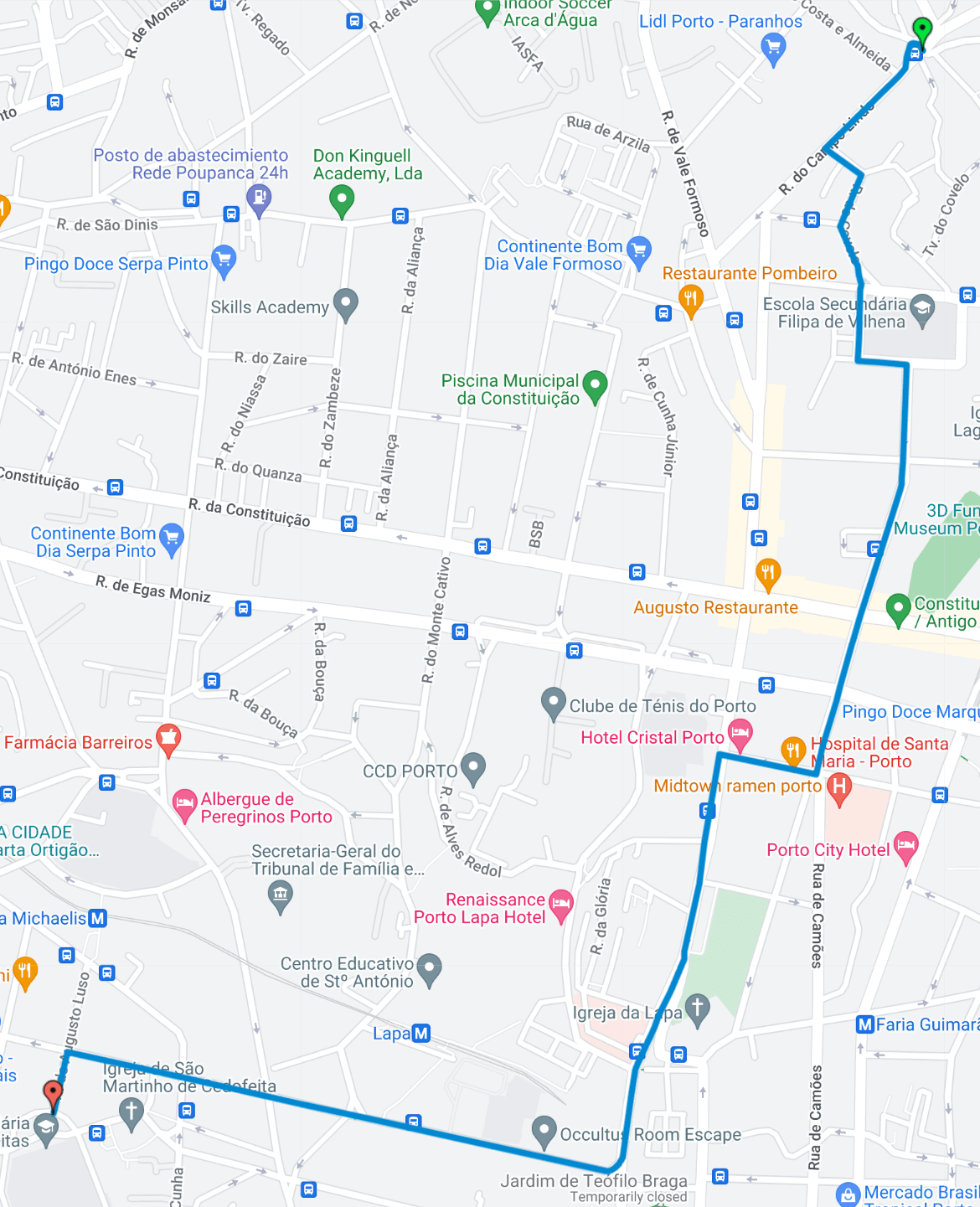}}
\caption{Visualization of the trajectory generated by HOSER and the real trajectory for the same OD pair in Porto.}
\label{fig:appendix_vis_traj_porto}
\end{figure*}

\begin{figure*}[t]
\centering
\subcaptionbox{Ours}{\includegraphics[width=0.4\textwidth]{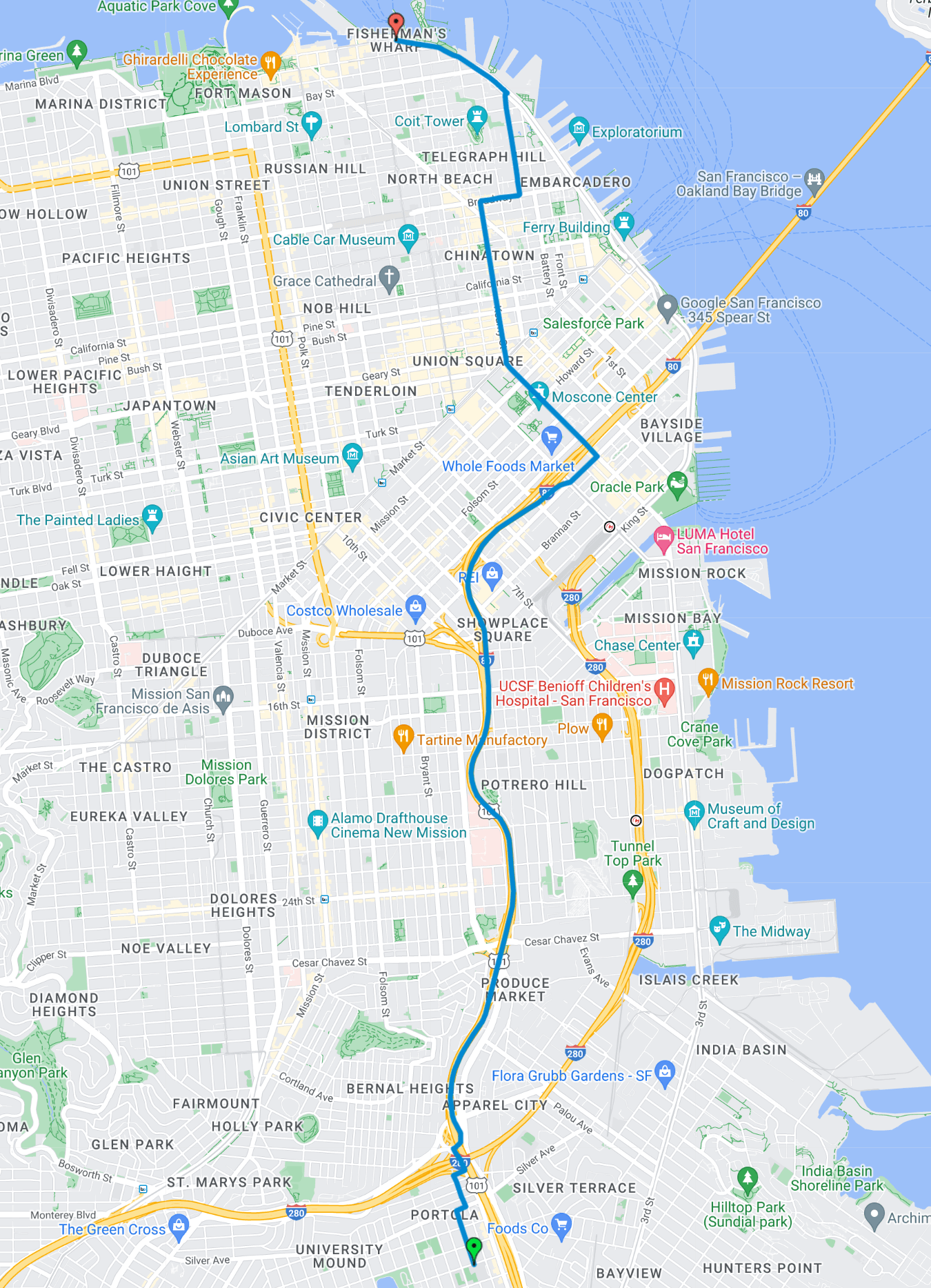}}
\subcaptionbox{Real}{\includegraphics[width=0.4\textwidth]{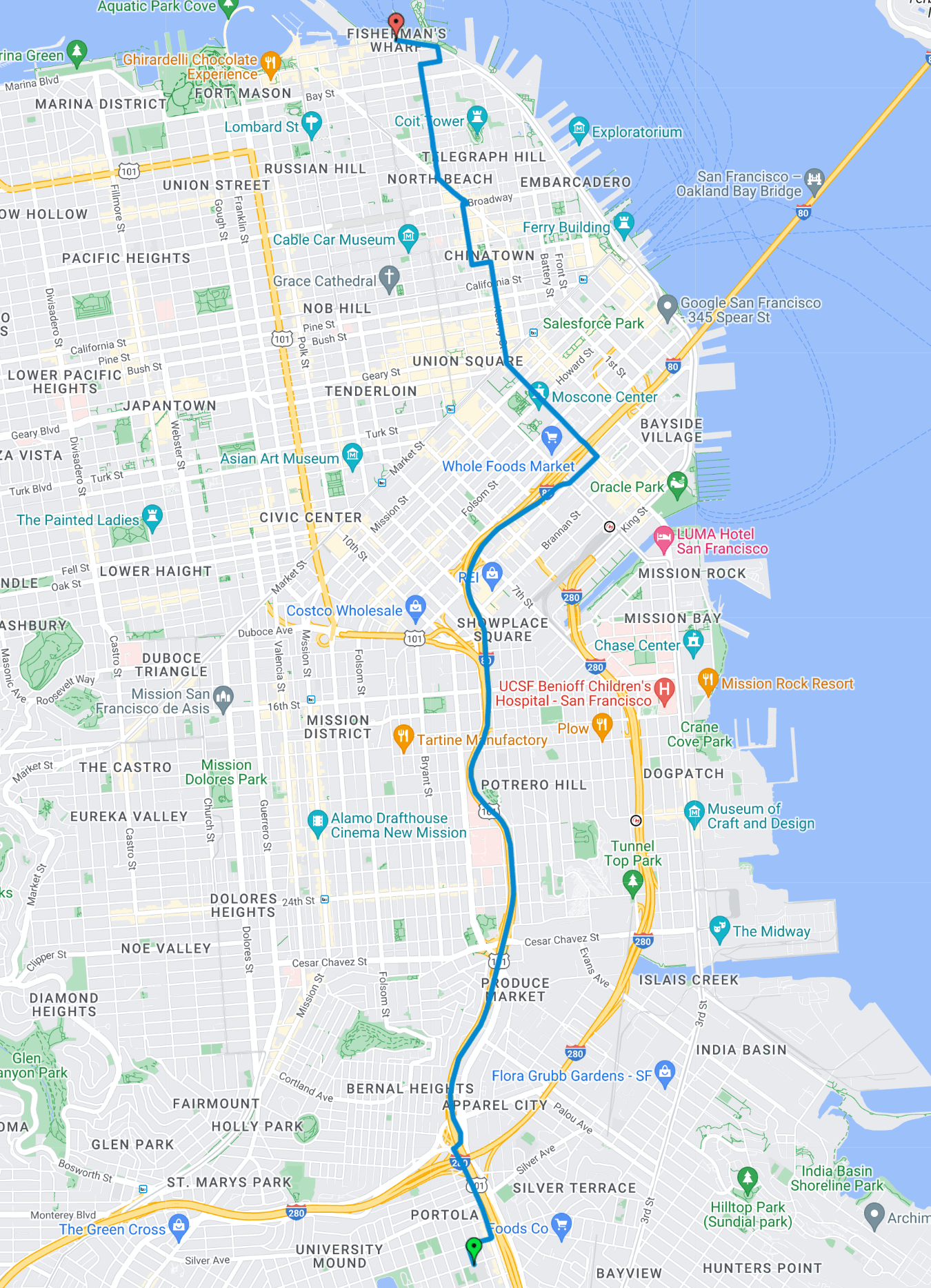}}
\caption{Visualization of the trajectory generated by HOSER and the real trajectory for the same OD pair in San Francisco.}
\label{fig:appendix_vis_traj_san_francisco}
\end{figure*}

Additionally, we visually compare the trajectories generated by HOSER with the real ones for the same OD pairs. The results, shown in \Cref{fig:appendix_vis_traj_beijing}, \Cref{fig:appendix_vis_traj_porto}, and \Cref{fig:appendix_vis_traj_san_francisco} for the three datasets, demonstrate that the generated trajectories closely match the real ones.

\subsection{Utility of Generated Data}

To evaluate the effectiveness of trajectories generated by various models in downstream tasks, we select the next-location prediction task as a benchmark for comparison. Specifically, we train the widely recognized next-location prediction model, DeepMove~\cite{feng2018deepmove}, using both real trajectories and generated trajectories. We then compare the impact of different training datasets on the model's performance, as illustrated in \Cref{tab:appendix_data_utility_comparison}. Notably, the trajectories generated by the SVAE, TSG, and DiffTraj models are based on GPS coordinates rather than road segments, requiring map-matching to ensure consistency with road network topology. This process, however, may introduce additional errors, particularly when the generated trajectories deviate significantly from the road network. As a result, we exclude these models from our comparative analysis. It can be observed that the trajectories generated by HOSER outperform all baselines in downstream tasks, demonstrating its superior ability to preserve the spatio-temporal characteristics of real-world trajectories and ensuring better applicability in downstream applications.

\begin{table}[ht]
\centering
\small
\setlength{\tabcolsep}{1mm}
\begin{tabular}{lcccccc}
\toprule
& \multicolumn{2}{c|}{Beijing} & \multicolumn{2}{c|}{Porto} & \multicolumn{2}{c}{San Francisco} \\
\cmidrule(lr){2-7}
Methods & Acc@5 & \multicolumn{1}{c|}{MRR} & Acc@5 & \multicolumn{1}{c|}{MRR} & Acc@5 & MRR \\
\midrule
Markov & 0.745 & \multicolumn{1}{c|}{0.657} & 0.846 & \multicolumn{1}{c|}{\underline{0.751}} & 0.761 & 0.612 \\
Dijkstra & \underline{0.746} & \multicolumn{1}{c|}{\underline{0.659}} & 0.841 & \multicolumn{1}{c|}{0.750} & \underline{0.763} & 0.647 \\
SeqGAN & 0.713 & \multicolumn{1}{c|}{0.637} & 0.795 & \multicolumn{1}{c|}{0.652} & 0.698 & 0.577 \\
SVAE & \ding{55} & \multicolumn{1}{c|}{\ding{55}} & \ding{55} & \multicolumn{1}{c|}{\ding{55}} & \ding{55} & \ding{55} \\
MoveSim & 0.662 & \multicolumn{1}{c|}{0.609} & 0.841 & \multicolumn{1}{c|}{0.679} & 0.665 & 0.539 \\
TSG & \ding{55} & \multicolumn{1}{c|}{\ding{55}} & \ding{55} & \multicolumn{1}{c|}{\ding{55}} & \ding{55} & \ding{55} \\
TrajGen & 0.224 & \multicolumn{1}{c|}{0.201} & 0.356 & \multicolumn{1}{c|}{0.304} & 0.253 & 0.229 \\
DiffTraj & \ding{55} & \multicolumn{1}{c|}{\ding{55}} & \ding{55} & \multicolumn{1}{c|}{\ding{55}} & \ding{55} & \ding{55} \\
STEGA & 0.736 & \multicolumn{1}{c|}{0.642} & 0.810 & \multicolumn{1}{c|}{0.668} & 0.757 & \underline{0.654} \\
TS-TrajGen & 0.743 & \multicolumn{1}{c|}{0.654} & \underline{0.857} & \multicolumn{1}{c|}{0.741} & 0.726 & 0.628 \\
\cmidrule(lr){1-7}
Ours & \textbf{0.776} & \multicolumn{1}{c|}{\textbf{0.697}} & \textbf{0.888} & \multicolumn{1}{c|}{\textbf{0.758}} & \textbf{0.797} & \textbf{0.673} \\
\cmidrule(lr){1-7}
Real & 0.804 & \multicolumn{1}{c|}{0.728} & 0.929 & \multicolumn{1}{c|}{0.780} & 0.847 & 0.698 \\
\bottomrule
\end{tabular}
\caption{Performance of DeepMove (a next-location prediction model) trained with different trajectory data. The best one is denoted by \textbf{boldface} and the second-best is denoted by \underline{underline}. Unsupported metrics are marked by \ding{55}. A higher value is preferable.}
\label{tab:appendix_data_utility_comparison}
\end{table}

\end{document}